\documentclass{article} 
\usepackage{iclr2027_conference,times}

\usepackage{hyperref}
\usepackage{url}

\usepackage{kotex}
\usepackage{booktabs}
\usepackage{graphicx}
\usepackage{amsmath,amssymb}
\usepackage{multirow}
\usepackage{pifont}

\title{BV Loss: Block Verification-Aware Loss \\ for Block Diffusion Speculative Decoding}

\DeclareRobustCommand{\authoremail}[1]{{\hypersetup{hidelinks}\href{mailto:#1}{\texttt{#1}}}}

\author{
\textbf{Suyoung Kim\textsuperscript{1,2}\thanks{First author. Email: \authoremail{ksyo96@snu.ac.kr}.},
Jahyun Koo\textsuperscript{1,2},
Hyeonjin Kim\textsuperscript{1,2},
Inhyeok Bang\textsuperscript{1},} \\
\textbf{Seunghyun Lee\textsuperscript{1},
Hyunjae Oh\textsuperscript{1,3},
Baeseong park\textsuperscript{1},
Dongsoo Lee\textsuperscript{1}\thanks{Corresponding author. Email: \authoremail{dongsoo.lee@a2sys.ai}.}} \\[0.5em]
{\hypersetup{hidelinks}\textsuperscript{1}a2sys (\href{https://a2sys.ai}{a2sys.ai})} \\
{\hypersetup{hidelinks}\textsuperscript{2}Seoul National University (\href{https://snu.ac.kr}{snu.ac.kr})} \\
{\hypersetup{hidelinks}\textsuperscript{3}University of Wisconsin (\href{https://wisc.edu}{wisc.edu})}
}

\newcommand{\hj}[1]{\textcolor{black}{#1}}
\newcommand{\mt}[1]{\textcolor{black}{#1}}

\iclrfinalcopy 
\begin{document}

\maketitle
\fancyhead{} 
\fancyhead[L]{Preprint.}

\begin{abstract}
Diffusion drafters accelerate speculative decoding by proposing multiple tokens in parallel.
Despite recent advances in speculative decoding through sequence-level drafting and verification, existing training objectives remain largely designed around token-level verification.
To address this mismatch, we introduce \textbf{Block Verification-aware loss (BV loss)}, a training objective designed to maximize the expected acceptance length of a drafted sequence.
BV loss is directly derived from the block verification acceptance rule, providing a principled connection between the drafter training objective and the inference-time verification mechanism at the sequence level. 
Across math, code, and chat benchmarks, BV loss increases the mean number of tokens accepted per verification call under block verification by 13.0--21.0\% over cross-entropy loss training for DFlash and DSpark with Qwen3-4B and Qwen3-8B without changing the inference procedure. 
BV loss also outperforms tokenwise acceptance objectives such as TV loss and LK loss, and its gains extend to token verification and greedy decoding. These results demonstrate the benefit of training block diffusion drafters with an objective aligned with sequence-level verification, rather than optimizing each token independently.

\end{abstract}

\section{Introduction}
\label{sec:introduction}

\suppressfloats[t]
\begin{table}[t]
 \centering
 \definecolor{comparisonYesColor}{HTML}{34786B}
 \definecolor{comparisonNoColor}{HTML}{B55353}
 \newcommand{\comparisonYes}{\textcolor{comparisonYesColor}{\ding{51}}}
 \newcommand{\comparisonNo}{\textcolor{comparisonNoColor}{\ding{55}}}
 \newcommand{\comparisonCell}[2]{\parbox[c]{#1}{\raggedright #2}}
\caption{\textbf{Comparison of drafter training objectives.}
CE and KL / RKL optimize token likelihood or distribution matching, whereas TV and LK loss explicitly target token-level acceptance.
BV loss is instead derived from block verification and directly optimizes the expected accepted draft length, making it both verification-aware and block-aware.}
 \label{tab:loss_positioning}
 \fontsize{10}{12}\selectfont
 \setlength{\tabcolsep}{2pt}
 \renewcommand{\arraystretch}{1.0}
 \begin{tabular*}{\linewidth}{@{\extracolsep{\fill}}llccc@{}}
  \toprule
  \comparisonCell{53pt}{\textbf{Drafter\\loss}}
   & \comparisonCell{153pt}{\textbf{Training objective}}
   & \parbox[c]{65pt}{\centering\bfseries Uses teacher\\probabilities?}
   & \parbox[c]{59pt}{\centering\bfseries Verification-\\aware?}
   & \parbox[c]{49pt}{\centering\bfseries Block-\\aware?}\\
  \midrule
  CE
   & \comparisonCell{153pt}{Maximize log-likelihood}
   & \comparisonNo & \comparisonNo & \comparisonNo\\[1pt]
  KL / RKL
   & \comparisonCell{153pt}{Minimize KL divergence}
   & \comparisonYes & \comparisonNo & \comparisonNo\\[1pt]
  TV
   & \comparisonCell{153pt}{Maximize token acceptance}
   & \comparisonYes & \comparisonYes & \comparisonNo\\[1pt]
  $\mathrm{LK}$
   & \comparisonCell{153pt}{Maximize token acceptance}
   & \comparisonYes & \comparisonYes & \comparisonNo\\[1pt]
  \midrule
  \textbf{BV (Ours)}
   & \comparisonCell{153pt}{\textbf{Maximize block acceptance length}}
   & \comparisonYes & \comparisonYes & \comparisonYes\\
  \bottomrule
 \end{tabular*}
\end{table}

Speculative decoding accelerates large language model (LLM) inference by \mt{drafting candidate tokens using a lightweight drafter model} and verifying them in parallel with a larger target model~\citep{leviathan2023speculative}. Its efficiency largely depends on how many drafted tokens \mt{are} accepted per \hj{verification}, making both drafting and verification critical to \mt{the} overall speedup. While the majority of \hj{related} research focuses on improving the drafter, another line of work seeks additional speedup through better verification algorithms. 
Block verification~\citep{sun2025block} is \mt{a key} approach that targets the verification side of speculative decoding. It introduces a sequence-level acceptance criterion and performs rejection sampling over the joint probabilities of candidate sequences. 
\mt{The} conventional verification scheme, which \mt{we call} token verification, terminates when a proposed token is rejected, causing all subsequent draft tokens to be discarded. By contrast, block verification evaluates the drafted sequence jointly and can \hj{accept tokens that would \mt{otherwise be} discarded under token verification}.
Importantly, replacing token verification with block verification incurs negligible additional inference overhead, while achieving an equal or higher acceptance rate.

Despite the \hj{emergence of} sequence-level verification in speculative decoding, existing drafters are still trained with token-level objectives.
A prominent example is total variation (TV) loss, which is derived from the single-token acceptance probability under token verification
Minimizing TV loss therefore provides a principled way to train a drafter for tokenwise rejection sampling.
However, under block verification, acceptance is determined \mt{by} the probability of the block-sized sequence, raising the question of whether token-level losses remain the appropriate training objectives.


\mt{To address this mismatch between the training objective and the verification criterion}, we propose \textbf{Block Verification-aware loss (BV loss)}, a drafter training objective designed to directly maximize the expected acceptance length of a drafted sequence under block verification. 
\mt{Our formulation is derived from the sequence-level acceptance criterion used in block verification rather than the token-level rejection sampling.} 
In this sense, BV loss serves as a sequence-level counterpart to TV loss: just as TV loss \mt{follows from} the acceptance probability induced by token-level rejection sampling, BV loss is derived from the acceptance length induced by block verification (see Table~\ref{tab:loss_positioning}).

We evaluate BV loss across multiple speculative-decoding configurations, drafter architectures, and target models. 
Drafters trained with BV loss consistently achieve longer accepted sequence lengths than those trained with conventional token-level objectives, demonstrating the benefit of aligning drafter training with sequence-level verification. 
Moreover, BV loss generalizes across different drafting and verification configurations. Drafters trained with BV loss outperform those trained with conventional objectives even under greedy decoding and token-level verification. These results show that BV loss is not only theoretically motivated by block verification, but also serves as an effective training objective across diverse speculative-decoding settings.

Our contributions are summarized below:
\begin{itemize}
    \item \textbf{A sequence-level objective aligned with block verification.}
    We introduce BV loss, a sequence-level training objective derived from the block verification acceptance rule. We prove that minimizing BV loss is equivalent to maximizing the expected accepted draft length. The objective accounts for both partial and full block acceptance and recovers TV loss as the single-token special case.

    \item \textbf{Theoretical analysis and practical optimization.}
    We show that the same tokenwise acceptance rate can yield different expected accepted lengths under block verification, exposing a limitation of tokenwise objectives. Our gradient analysis shows how low acceptance at earlier positions weakens learning signals at later positions, motivating a logarithmic objective that preserves the optimum while strengthening optimization in low-acceptance regimes. We further derive blockwise training surrogates from target-generated trajectories, enabling training from scratch without differentiating through draft sampling.

    \item \textbf{Consistent gains across drafters and verification rules.}
    Across seven math, code, and chat benchmarks, BV loss increases
    the mean number of tokens retained per verification call by
    13.0--21.0\% over CE training for DFlash and DSpark with Qwen3-4B
    and Qwen3-8B under block verification. These gains extend to
    token verification and greedy decoding without changing the
    inference procedure.
\end{itemize}
\section{Related Work}
\label{sec:related_work}

\begin{figure}[t]
 \centering
 \includegraphics[width=\linewidth]{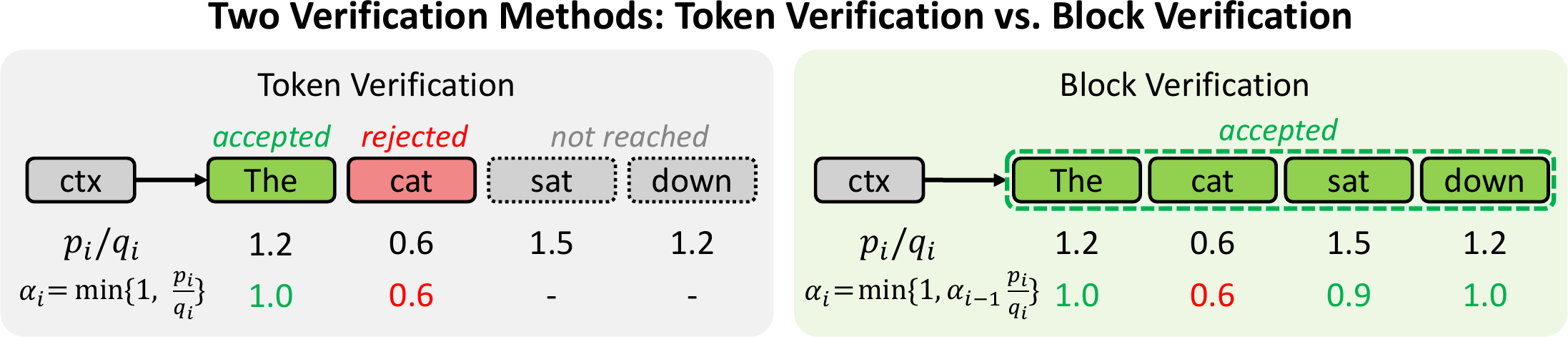}
    \caption{\textbf{Token Verification versus Block Verification.}
    Given the same sequence of target-to-draft probability ratios $p_i/q_i$,
    Token Verification accepts tokens sequentially and terminates at the first
    rejection; in this example, \texttt{cat} is rejected and the remaining
    tokens are never reached.
    Block Verification instead propagates the acceptance state across positions,
    $\alpha_i=\min\{1,\alpha_{i-1}p_i/q_i\}$, allowing a low ratio at one
    position to be compensated by sufficiently large ratios at subsequent
    positions.
    As a result, the entire draft block can be retained even when Token
    Verification would stop early.}
 \label{fig:verifications}
\end{figure}

\paragraph{Speculative Decoding.}
Speculative decoding accelerates LLM inference by using a lightweight drafter to propose candidate tokens, while verification and correction by the target model preserve the target distribution and ensure lossless generation~\citep{leviathan2023speculative,chen2023speculative}.
Early approaches primarily relied on autoregressive drafters that generated candidate tokens sequentially~\citep{leviathan2023speculative,eagle,li2024eagle2,li2025eagle3}. 
More recent work has explored parallel drafting to reduce the sequential overhead of proposal generation. DFlash~\citep{chen2026dflash} employs a block diffusion model to generate draft tokens in parallel, while DSpark~\citep{cheng2026dspark} introduces causal correction to capture dependencies among draft tokens.
Our work is complementary to these architectural advances: rather than modifying the drafter architecture, we study how diffusion drafters should be trained when their proposals are evaluated by block verification.

\paragraph{Verification for Speculative Decoding.}
Exact verification in speculative decoding adapts rejection sampling to correct drafter proposals while ensuring that the resulting output distribution \mt{is identical to} that of the target model. 
Figure~\ref{fig:verifications} illustrates how block verification differs from the token verification.
In this illustration, $p_i/q_i$ is the ratio of target and draft probabilities, while $\alpha_i$ denotes the acceptance probability. In the shown example, both methods verify the same draft: token verification could stop at the second token, where \(p_i/q_i=0.6\), whereas block verification retains the full block, yielding a longer accepted draft length.
Our work focuses on the corresponding training problem: existing drafter objectives optimize tokens independently, whereas block verification determines acceptance jointly at the sequence level.

\begin{figure}[ht]
 \centering
 \includegraphics[width=\linewidth]{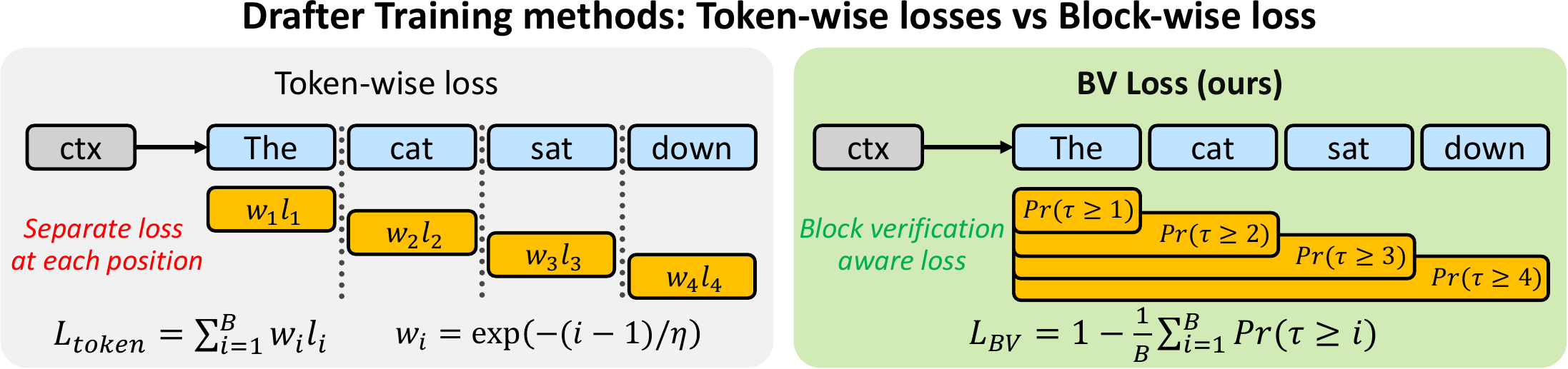}
 \caption{\textbf{Token-wise losses versus BV Loss.}
Token-wise objectives optimize each draft position separately, typically
through weighted local losses $w_i\ell_i$.
BV Loss instead couples positions through the prefix-retention
probabilities $\Pr(\tau \ge i)$ under Block Verification.
Since $\sum_{i=1}^{B}\Pr(\tau \ge i)$ equals the expected number of
accepted draft tokens, BV Loss directly optimizes the quantity of interest
at verification time.
}
 \label{fig:overview}
\end{figure}

\paragraph{Verification-aware Training Objectives.}
Total Variation loss (TV loss) serves as a direct objective for the per-token acceptance rate, with its formulation derived from the rejection-sampling criterion used in token verification. 
\mt{Consequently, minimizing TV loss optimizes the expected
token-level acceptance rate of the drafter under tokenwise verification.} 
However, optimizing TV loss alone often leads to unstable or ineffective training. Building on this observation, LK losses \mt{\citep{samarin2026lk}} propose a practical objective based on the negative logarithm of the token-level acceptance probability. 
While such objectives are well aligned with token verification, their advantage of optimizing the relevant acceptance rate does not naturally extend to block verification. This mismatch motivates a training objective derived from the sequence-level acceptance criterion used by block verification (see~\ref{fig:overview}).



\section{Method}
\label{sec:method}

\subsection{Preliminaries: Tokenwise Draft Training}
\label{sec:preliminaries}

\paragraph{\hj{Notations.}}
A drafter proposes $B$ tokens, which the target verifies in one call. For a target-generated training block $y_1,\ldots,y_B$, let $p_i(v)$ and $q_i(v)$ denote the target and draft probabilities of token $v$ at position $i$, evaluated on the same prefix. The target is frozen. We omit the shared context and model parameters from the notation.

\paragraph{\hj{Tokenwise Optimization.}}

Common per-token loss functions include cross-entropy (CE), Kullback–Leibler divergence (KL), reverse KL divergence (RKL), and total variation (TV):

\begin{equation}
 \begin{aligned}
 \ell_i^{\mathrm{CE}}&=-\log q_i(y_i),
 &\ell_i^{\mathrm{KL}}&=\textstyle\sum_v p_i(v)\log\tfrac{p_i(v)}{q_i(v)},\\[4pt]
 \ell_i^{\mathrm{RKL}}&=\textstyle\sum_v q_i(v)\log\tfrac{q_i(v)}{p_i(v)},
 &\ell_i^{\mathrm{TV}}&=\tfrac12\textstyle\sum_v|p_i(v)-q_i(v)|.
 \end{aligned}
 \label{eq:token_losses}
\end{equation}

For a drafter with block size $B$, per-position loss $\ell_i$ is aggregated over $i=1,\ldots,B$ as a weighted sum. A common training heuristic gives earlier positions more weight because later tokens are useful only if the earlier prefix is retained. For example, DFlash~\citep{chen2026dflash} applies exponential position decay to its cross-entropy loss:

\begin{equation}
 \mathcal L_{\mathrm{token}} \triangleq \sum_{i=1}^{B}w_i\ell_i,
 \qquad w_i=\exp\left(-\frac{i-1}{\eta}\right),\qquad\eta>0.
 \label{eq:token_weighting}
\end{equation}

Smaller $\eta$ concentrates training more on earlier positions \hj{and less on later positions}. 
We refer to training with $\mathcal{L}_\mathrm{token}$ as tokenwise optimization.

\paragraph{\hj{Token Verification Awareness.}}

\hj{With tokenwise rejection sampling of token verification}, a candidate token $v$ at position $i$, drawn from $q_i$ is accepted with probability $\alpha_i =  \min\{1,p_i(v)/q_i(v)\}$. Averaging over the candidates \hj{at the same position gives the expected acceptance rate $\mathbb{E}_\mathrm{TV}[\alpha_i]$}:

\begin{equation}
  \mathbb{E}_\mathrm{TV}[\alpha_i]
 =\sum_v q_i(v)\min\left\{1,\frac{p_i(v)}{q_i(v)}\right\}
 =\sum_v\min\{p_i(v),q_i(v)\}.
 \label{eq:tv_acceptance}
\end{equation}

\hj{From $\min\{p,q\}=(p+q-|p-q|)/2$, it follows that $\mathbb{E}_\mathrm{TV}[\alpha_i]=1-\ell_i^{\mathrm{TV}}$.} Thus minimizing TV loss \hj{$\ell_i^\mathrm{TV}$ directly} maximizes single-token acceptance. However, its gradients can be very small at random initialization, \hj{making pure TV optimization impractical.} \mt{LK losses}~ \citep{samarin2026lk} address this difficulty using LK loss:

\begin{equation}
 \begin{aligned}
 \ell^{\mathrm{LK}, \alpha}_i
   &=-\log\mathbb{E}_\mathrm{TV}[\alpha_i]=-\log(1-\ell_i^{\mathrm{TV}}).
 \end{aligned}
 \label{eq:lk_losses}
\end{equation}

$\ell^{\mathrm{LK}, \alpha}_i$ minimizes the negative log value of the per-position expected acceptance rate $\mathbb{E}_\mathrm{TV}[\alpha_i]$. 
While TV loss and its variant LK loss have their advantage of directly optimizing token verification, their effectiveness becomes questionable when used with other verification methods.


\subsection{Block Verification-aware Loss}
\label{sec:bv_identity}

\hj{In this section, we present \textbf{Block Verification-aware loss (BV loss)}, a drafter training objective that is designed to directly optimize the acceptance rate under block verification. The derivation of BV loss is introduced step-by-step in the following paragraphs.}

\paragraph{\hj{Block Verification Awareness.}}
As with \hj{token verification}, we start from the \hj{expected acceptance rate at each prefix length of a drafted block $y_{1:B}\sim q$}.
\hj{The conditional probability of retaining the first $i$ drafted tokens is defined from the block-verification acceptance rule of \citet{sun2025block} as}:

\begin{equation}
 \alpha_i(y_{1:i}) = \min\left\{1,\alpha_{i-1}(y_{1:i-1})\frac{p_i(y_i)}{q_i(y_i)}\right\},
 \qquad \alpha_0=1.
 \label{eq:block_acceptance}
\end{equation}

Here, $\alpha_i(y_{1:i})$ is the acceptance probability of the length-$i$ drafted prefix, and $y_i$ is the $i$-th token in the drafted sequence block $y_{1:B}\sim q$.  \hj{Let $\tau$ denote the number of accepted draft tokens. Averaging over drafted blocks gives the unconditional prefix acceptance rate: $\mathbb E_{\mathrm{BV}}[\alpha_i] \triangleq \mathbb E_{y_{1:B}\sim q}[\alpha_i(y_{1:i})]=\Pr_{\mathrm{BV}}(\tau\ge i)$. Then the expected acceptance length is given by:}

\begin{equation}
 \mathbb{E}_{\mathrm{BV}}[\tau]
  =\sum_{i=1}^{B}\Pr\nolimits_{\scriptscriptstyle\mathrm{BV}}(\tau\ge i)
  =\sum_{i=1}^{B}\mathbb{E}_\mathrm{BV}[\alpha_i].
 \label{eq:prefix_identity}
\end{equation}

\paragraph{\hj{Draft Acceptance Approximation.}}

\hj{While directly evaluating $\mathbb{E}_{\mathrm{BV}}[\alpha_i]$ requires draft-generated prefixes $y_{1:i}\sim q$, the frozen-target training data provide prefixes $y^\prime_{1:i}\sim p$. We therefore derive a form of Eq.~\eqref{eq:block_acceptance}  which permits estimating Eq.~\eqref{eq:prefix_identity} on target-generated prefixes for training.} Define the \hj{cumulative ratio through position $i$}:

\begin{equation}
 r_i\triangleq\prod_{j=1}^{i}\frac{q_j(y^\prime_j)}{p_j(y^\prime_j)},
       \qquad r_0=1,
 \label{eq:prefix_states}
\end{equation}

\hj{with the assumption that training paths are sampled from the target conditionals $p_i$ used in the ratios, $q_i$ is the proposal conditional used at inference, and $p_i$ has support wherever $q_i$ is nonzero.}
\hj{Multiplying $\alpha_i(y^\prime_{1:i})$ by $r_i$ yields the acceptance score $\widetilde{\alpha}_i$:}

\begin{equation}
 \widetilde{\alpha}_i(y^\prime_{1:i})\triangleq r_i \alpha_i(y^\prime_{1:i})
     =\min\{r_i, \widetilde{\alpha}_{i-1}(y^\prime_{1:i-1})\}
     =\min\{1,r_1,\ldots,r_i\}, \qquad \widetilde{\alpha}_0=1.
     \label{eq:acceptance_score}
\end{equation}

\hj{Assume that training paths are sampled from the target conditionals $p_i$ used in the ratios and that $q_i$ is the proposal conditional used at inference. Under these conditions, $\mathbb{E}_{y^\prime_{1:B}\sim p}[\widetilde{\alpha}_i] =\mathbb{E}_{\mathrm{BV}}[\alpha_i]$ for every $i$, and summing over positions yields Eq.~\eqref{eq:target_prefix_identity}:}

\begin{equation}
 \mathbb{E}_{y^\prime_{1:B}\sim p}
 \left[\sum_{i=1}^{B}\widetilde{\alpha}_i(y^\prime_{1:i})\right]
 =
 \mathbb{E}_{\mathrm{BV}}[\tau].
 \label{eq:target_prefix_identity}
\end{equation}

This identity also holds after averaging over any fixed distribution of training contexts.

\paragraph{Block Verification-aware Loss.}

\hj{To this end, we define Block Verification-aware loss (BV loss) as one minus the expected acceptance length normalized by the block size $B$}:




\begin{equation}
 \mathcal L_{\mathrm{BV}}\triangleq 1-\frac{1}{B}\mathbb{E}_{\mathrm{BV}}[\tau].
 \label{eq:linear_bv}
\end{equation}

Using Eq.~\eqref{eq:target_prefix_identity}, it can be evaluated on target-generated training paths as
\begin{equation}
 \mathcal L_{\mathrm{BV}}
 =
 1-\frac{1}{B}\mathbb E_{y^\prime_{1:B}\sim p}
 \left[\sum_{i=1}^{B}\widetilde{\alpha}_i(y^\prime_{1:i})\right].
 \label{eq:linear_bv_train}
\end{equation}

\hj{Minibatch training estimates the objective by averaging the per-block quantity across target-generated blocks. Minimizing \mt{the BV loss} is exactly equivalent to maximizing expected BV acceptance length. For $B=1$, BV loss reduces to TV loss $\ell^\mathrm{TV}$.}

The cumulative ratios make BV loss a block objective. For example, for draft-to-target ratios $(1.25,0.8)$, block verification gives $\widetilde{\alpha}_2=\min(1,1.25,1)=1$, whereas independently clipping the ratios gives $\min\{1,1.25\}\min\{1,0.8\}=0.8$. A preceding likelihood surplus can therefore \mt{offset} a later deficit. \mt{As a result,} BV depends on how probability mass is arranged across prefixes, which the individual tokenwise overlaps do not capture.

\subsection{Gradient Analysis and the Log BV Loss}
\label{sec:gradients}
\label{sec:log_loss}

\paragraph{Gradients of the accepted-length objective.}
Let $n_i$ be the earliest position attaining the minimum in $\widetilde{\alpha}_i$,
including position zero. Away from ties,
\begin{equation}
 \nabla \widetilde{\alpha}_i
 =\widetilde{\alpha}_i\sum_{j=1}^{n_i}\nabla\log q_j(y_j),
 \qquad
 \nabla\mathbb{E}_{\mathrm{BV}}[\tau]
 =\mathbb{E}_{y\sim p}\left[\sum_i\nabla \widetilde{\alpha}_i\right].
 \label{eq:acceptance_gradient}
\end{equation}
The derivative accounts for every prefix whose active minimum depends
on the prediction. Its magnitude is scaled by the prefix scores $\widetilde{\alpha}_i$
and can be small when acceptance is low. We therefore normalize by the
total expected accepted length.

\paragraph{Log-scaled BV objective.}
We define the log BV objective as
\begin{equation}
 \mathcal L_{\mathrm{BV}}^{\log}
 =\log B-\log\mathbb{E}_{\mathrm{BV}}[\tau]
 \label{eq:population_bv_loss}
\end{equation}
The constant $\log B$ has zero gradient and is included only to keep the loss nonnegative. Since $-\log$ is strictly decreasing,
Eq.~\eqref{eq:population_bv_loss} preserves the \mt{minimizers
of Eq.~\eqref{eq:linear_bv} in expectation.}

Its gradient is
\begin{equation}
 \nabla\mathcal L_{\mathrm{BV}}^{\log}
 =-\frac{\nabla\mathbb{E}_{\mathrm{BV}}[\tau]}{\mathbb{E}_{\mathrm{BV}}[\tau]}
 =-\mathbb{E}_{y\sim p}\left[
   \sum_i\frac{\widetilde{\alpha}_i}{\mathbb{E}_{\mathrm{BV}}[\tau]}
   \sum_{j=1}^{n_i}\nabla\log q_j(y_j)\right].
 \label{eq:log_gradient}
\end{equation}
The prefix weights sum to one in expectation. Log scaling removes the
overall small-acceptance factor while preserving the direction of
expected-length improvement and \mt{the} prefix dependencies.

\paragraph{Practical training setup.}
We use an initial warm-up for stable training.
Log normalization controls the overall gradient scale, but prefixes with
small acceptance scores can still receive a negligible share of the update
at initialization. To accelerate early learning of these prefixes, we anneal the aggregation of BV scores. For positive sampled-label
scores, define the annealed BV objective as the negative log of their
power mean:
\begin{equation}
 \mathcal L_{\mathrm{BV}}^{\mathrm{anneal}}(\beta)
 =-\log\left[\left(\frac1B\sum_{i=1}^{B}
   \widetilde{\alpha}_i^\beta\right)^{1/\beta}\right],
 \qquad 0<\beta\le1.
 \label{eq:annealing}
\end{equation}
This mean interpolates between the geometric mean as $\beta\to0$ and
the arithmetic mean at $\beta=1$. Defining the loss at $\beta=0$ by
continuity gives the endpoints
\[
 \mathcal L_{\mathrm{BV}}^{\mathrm{anneal}}(0)
 =\frac1B\sum_{i=1}^{B}\bigl(-\log\widetilde{\alpha}_i\bigr),
 \qquad
 \mathcal L_{\mathrm{BV}}^{\mathrm{anneal}}(1)
 =-\log\left(\frac1B\sum_{i=1}^{B}\widetilde{\alpha}_i\right).
\]
Thus, annealing moves from averaging per-prefix log losses to taking
the log loss of the average prefix score. The latter is the blockwise
log BV surrogate; averaging it over blocks differs from taking the
logarithm of expected length in Eq.~\eqref{eq:population_bv_loss}.
Its gradient is a weighted average of per-prefix log-loss gradients:
\begin{equation}
 \nabla\mathcal L_{\mathrm{BV}}^{\mathrm{anneal}}(\beta)
 =\sum_{i=1}^{B}
   \frac{\widetilde{\alpha}_i^\beta}
        {\sum_{j=1}^{B}\widetilde{\alpha}_j^\beta}
   \nabla\!\left(-\log\widetilde{\alpha}_i\right).
 \label{eq:annealing_gradient}
\end{equation}
At $\beta=0$, every prefix contributes with coefficient $1/B$.
At $\beta=1$, its coefficient is proportional to its acceptance score.
For a score below the block mean, the initial coefficient is therefore
larger by a factor of $\sum_j\widetilde{\alpha}_j/(B\widetilde{\alpha}_i)$.
This strengthens the relative \mt{learning signal for low-acceptance prefixes}
and can speed up their early improvement before returning to the block-log
objective. In the main experiments, we increase $\beta$ linearly from 0 to 1
over the first three epochs, then fix it at 1; implementation details are
given in Appendix~\ref{sec:annealing_details}.

\section{Experiments}
\label{sec:experiments}

\subsection{Experimental Setup}
\label{sec:experimental_setup}

\paragraph{Baselines and training setups.}
We train five-layer DFlash and DSpark drafters with block size $B=15$ proposals from scratch using frozen Qwen3-4B and Qwen3-8B target models. All methods are trained for six epochs on the 100K target-generated responses generated from Open-PerfectBlend which is an open-sourced version of perfectBlend~\citep{xu2024perfect}. Table~\ref{tab:main_results} compares six training objectives introduced in Section~\ref{sec:preliminaries}: CE, KL, RKL, TV, LK, and BV. All experiments reported as BV use the log BV objective in Eq.~\eqref{eq:annealing}. For brevity, we refer to this objective simply as BV throughout the experimental section. Appendix~\ref{sec:appendix_implementation} provides further implementation details.

\paragraph{Benchmarks and evaluation.}
Following the evaluation protocol of previous work~\citep{chen2026dflash}, we evaluate across three task categories: \textbf{Math} with GSM8K~\citep{cobbe2021gsm8k}, MATH-500~\citep{lightman2024math500}, and AIME25~\citep{aime25}; \textbf{Code} with HumanEval~\citep{chen2021humaneval}, MBPP~\citep{austin2021mbpp}, and LiveCodeBench~\citep{jain2025livecodebench}; and \textbf{Chat} with MT-Bench~\citep{zheng2023mtbench}. We evaluate both greedy decoding ($T=0$) and sampling ($T=1$) under block verification. For each benchmark, we compute the average acceptance length $\tau$ as the total number of accepted tokens divided by the number of verification calls, including correction or bonus tokens when applicable. Speedup is the ratio of speculative to target-only autoregressive generation throughput (tokens/s), measured by an NVIDIA B200 GPU using FlashAttention-2~\citep{dao2024flashattention}. Table~\ref{tab:main_results} reports the average performance across the seven benchmarks.

\subsection{Main Results}
\label{sec:main_results}


\begin{table}[t]
\centering
\caption{Average performance over seven benchmarks on Qwen3-4B and Qwen3-8B
under greedy decoding ($T=0$) and block verification ($T=1$).
Average acceptance length $\tau$ includes correction/bonus tokens.
Speedup is relative to autoregressive decoding.
Boldface indicates the best average in each column within each target.}
\label{tab:main_results}
\small
\begin{tabular}{@{}ll*{8}{c}@{}}
\toprule
 & & \multicolumn{4}{c}{$T=0$ (greedy)}
 & \multicolumn{4}{c}{$T=1$ (block verification)}\\
\cmidrule(lr){3-6}\cmidrule(lr){7-10}
 & & \multicolumn{2}{c}{DFlash} & \multicolumn{2}{c}{DSpark}
 & \multicolumn{2}{c}{DFlash} & \multicolumn{2}{c}{DSpark}\\
\cmidrule(lr){3-4}\cmidrule(lr){5-6}\cmidrule(lr){7-8}\cmidrule(lr){9-10}
Target & Loss & $\tau$ & Speedup & $\tau$ & Speedup
 & $\tau$ & Speedup & $\tau$ & Speedup\\
\midrule
\multirow{6}{*}{Qwen3-4B} & CE & 4.65 & 3.71\ensuremath{\times} & 5.21 & 3.97\ensuremath{\times} & 4.09 & 3.21\ensuremath{\times} & 4.77 & 3.47\ensuremath{\times} \\
 & KL & 4.80 & 3.76\ensuremath{\times} & 5.49 & 4.15\ensuremath{\times} & 4.17 & 3.23\ensuremath{\times} & 4.97 & 3.67\ensuremath{\times} \\
 & RKL & 4.05 & 3.24\ensuremath{\times} & 4.87 & 3.73\ensuremath{\times} & 3.87 & 3.09\ensuremath{\times} & 4.68 & 3.45\ensuremath{\times} \\
 & TV & 1.98 & 1.60\ensuremath{\times} & 2.23 & 1.74\ensuremath{\times} & 1.96 & 1.57\ensuremath{\times} & 2.21 & 1.66\ensuremath{\times} \\
 & LK & 4.79 & 3.75\ensuremath{\times} & 5.54 & 4.24\ensuremath{\times} & 4.25 & 3.34\ensuremath{\times} & 5.15 & 3.77\ensuremath{\times} \\
 & \textbf{BV (Ours)} & \textbf{5.22} & \textbf{4.01\ensuremath{\times}} & \textbf{5.62} & \textbf{4.28\ensuremath{\times}} & \textbf{4.95} & \textbf{3.84\ensuremath{\times}} & \textbf{5.39} & \textbf{3.88\ensuremath{\times}} \\
\midrule
\multirow{6}{*}{Qwen3-8B} & CE & 4.64 & 3.60\ensuremath{\times} & 5.10 & 3.82\ensuremath{\times} & 4.15 & 3.29\ensuremath{\times} & 4.71 & 3.48\ensuremath{\times} \\
 & KL & 4.81 & 3.71\ensuremath{\times} & 5.58 & 4.09\ensuremath{\times} & 4.19 & 3.36\ensuremath{\times} & 5.07 & 3.76\ensuremath{\times} \\
 & RKL & 3.10 & 2.42\ensuremath{\times} & 3.50 & 2.69\ensuremath{\times} & 3.02 & 2.44\ensuremath{\times} & 3.40 & 2.52\ensuremath{\times} \\
 & TV & 1.01 & 0.81\ensuremath{\times} & 1.02 & 0.80\ensuremath{\times} & 1.01 & 0.81\ensuremath{\times} & 1.02 & 0.77\ensuremath{\times} \\
 & LK & 4.82 & 3.71\ensuremath{\times} & 5.57 & 4.18\ensuremath{\times} & 4.28 & 3.38\ensuremath{\times} & 5.20 & 3.82\ensuremath{\times} \\
 & \textbf{BV (Ours)} & \textbf{5.27} & \textbf{4.15\ensuremath{\times}} & \textbf{5.67} & \textbf{4.38\ensuremath{\times}} & \textbf{4.98} & \textbf{3.88\ensuremath{\times}} & \textbf{5.43} & \textbf{4.00\ensuremath{\times}} \\
\bottomrule
\end{tabular}
\end{table}

\paragraph{Consistent gains across target models and drafter methods.}
BV achieves the highest average acceptance length $\tau$ in every target--drafter configuration in Table~\ref{tab:main_results}. Under block verification at $T=1$, it improves over CE on all seven benchmarks for both drafters and both targets.
On Qwen3-4B, DFlash improves from 4.09 to 4.95
(\textbf{+21.0\%}), and DSpark from 4.77 to 5.39
(\textbf{+13.0\%}). On Qwen3-8B, the corresponding gains are
4.15 to 4.98 (\textbf{+20.0\%}) and 4.71 to 5.43
(\textbf{+15.3\%}). These results support the motivation of
Section~\ref{sec:bv_identity}: training for sequence-level acceptance
improves the number of tokens retained by a block verifier.
Appendix~\ref{sec:full_results} reports the benchmark-level results.

\paragraph{Beyond token acceptance optimization.}
LK already addresses weak TV gradients through log scaling, but still
optimizes acceptance separately at each position. For DFlash on Qwen3-4B,
BV increases mean $\tau$ under block verification from LK's 4.25 to 4.95
(\textbf{+16.5\%}) with the same training data and epoch budget.
The same comparison on Qwen3-8B improves from 4.28 to 4.98
(\textbf{+16.4\%}). DSpark also improves over LK on both targets
(5.15 to 5.39 and 5.20 to 5.43).
The gains over LK support extending acceptance-aware training from
individual tokens to draft prefixes, beyond the benefit of log scaling alone.

\paragraph{Decoding speed and greedy evaluation.}
For DFlash, the block-verification gains translate into higher measured
speedups: $3.21\times$ to $3.84\times$ on Qwen3-4B and
$3.29\times$ to $3.88\times$ on Qwen3-8B.
DSpark improves from $3.47\times$ to $3.88\times$ on Qwen3-4B
and from $3.48\times$ to $4.00\times$ on Qwen3-8B.
BV also improves $\tau$ over CE on every benchmark under greedy decoding,
with mean gains of 12.3\% and 7.9\% for DFlash and DSpark on Qwen3-4B,
and 13.6\% and 11.2\%, respectively, on Qwen3-8B.

\subsection{Analysis and Ablation}
\label{sec:analysis_ablation}

\begin{figure}[t]
\centering
\includegraphics[width=\linewidth]{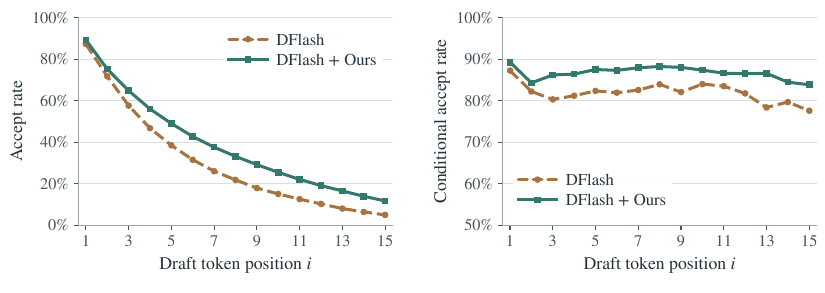}
\vspace{-1cm}
\caption{\textbf{Position-wise draft retention under block verification.}
We compare DFlash and DFlash + Ours on 512 GSM8K prompts
using Qwen3-4B.
\textbf{Left:} Unconditional prefix retention,
$\Pr_{\mathrm{BV}}(\tau \ge i)$, measuring the probability that the draft
is retained through position $i$.
\textbf{Right:} Conditional prefix retention,
$\Pr_{\mathrm{BV}}(\tau \ge i \mid \tau \ge i-1)$, measuring the probability
of extending an already retained prefix by one additional token.}
\label{fig:position_acceptance}
\end{figure}
\begin{figure}[t]
\centering
\includegraphics[width=\linewidth]{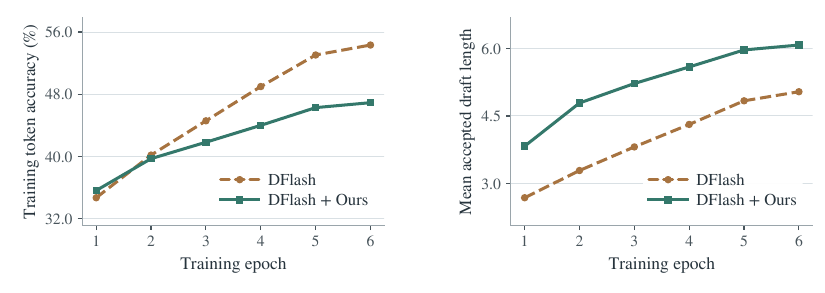}
\vspace{-1cm}
\caption{\textbf{Token accuracy and block acceptance exhibit different
training dynamics.}
We evaluate DFlash and DFlash + Ours on a fixed subset of 512
Qwen3-4B training examples.
\textbf{Left:} Token accuracy measured by argmax agreement with stored
target tokens.
\textbf{Right:} Mean accepted draft length under block verification.
While DFlash (CE) eventually achieves higher training token accuracy, BV consistently
produces longer accepted drafts, highlighting the distinction between
token-level prediction quality and sequence-level acceptance.
Accepted length excludes anchor and correction/bonus tokens.}
\label{fig:training_dynamics}
\end{figure}

\paragraph{Acceptance across positions.}
Figure~\ref{fig:position_acceptance} compares the final DFlash CE and BV
checkpoints on 512 GSM8K prompts with Qwen3-4B and block verification. The left panel measures unconditional prefix retention,
$\Pr_{\mathrm{BV}}(\tau\ge i)=\mathbb{E}_{\mathrm{BV}}[\alpha_i]$.
First-position retention is similar (87.2\% for CE and 89.2\% for BV),
while retention at position 8 increases from 21.8\% to 33.2\%.
By Eq.~\eqref{eq:prefix_identity}, the sum of these probabilities equals
the mean accepted draft length, which rises from 4.56 to 5.86
(\textbf{+28.4\%}). The right panel measures the probability of extending an
already retained prefix. BV improves this rate at all 15 positions,
including an increase from 77.6\% to 83.8\% at the final position.
Complete-block retention more than doubles, from 4.97\% to 11.69\%.
The improvement therefore extends beyond the first token to the longer
prefixes credited by BV loss.

\paragraph{Training dynamics.}
Figure~\ref{fig:training_dynamics} compares token accuracy and accepted draft
length on a fixed subset of 512 training examples, using eight shared
blocks per example across all six epochs. Here, accuracy measures argmax
agreement with sampled target tokens.
At epoch six, CE matches 54.33\% of the stored target tokens, compared with
46.91\% for BV, while BV retains 20.6\% more draft tokens
(6.07 versus 5.04). The reversed ranking shows that token accuracy is an
incomplete proxy for block acceptance, motivating optimization with the block verification-aware training.

\paragraph{Gains extend to token verification.}
Table~\ref{tab:token_verification} evaluates the same checkpoints with token
verification at $T=1$. BV achieves the highest mean $\tau$ in all target--drafter configurations, improving over CE by
20.9\% and 12.1\% for DFlash and DSpark on Qwen3-4B,
and by 20.3\% and 13.6\% on Qwen3-8B. Although BV is
derived from block verification, these results show that the learned improvements also benefit token verification without retraining.

\begin{table}[t]
\centering
\caption{Generalization to token verification at $T=1$ with the same checkpoints as
Table~\ref{tab:main_results}. Entries are seven-benchmark mean $\tau$, using the
same counting convention. Bold marks the best mean in each row, including ties.}
\label{tab:token_verification}
\renewcommand{\arraystretch}{1.10}
\begin{tabular}{@{}ll*{6}{c}@{}}
\toprule
\multirow{2}{*}{Target} & \multirow{2}{*}{Drafter}
 & \multicolumn{6}{c}{$\tau$ under token verification}\\
\cmidrule(lr){3-8}
 & & CE & KL & RKL & TV & LK
 & \textbf{BV (Ours)}\\
\midrule
\multirow{2}{*}{Qwen3-4B} & DFlash & 4.01 & 4.11 & 3.86 & 1.96 & 4.17 & \textbf{4.85} \\
 & DSpark & 4.64 & 4.84 & 4.64 & 2.20 & 4.98 & \textbf{5.20} \\
\midrule
\multirow{2}{*}{Qwen3-8B} & DFlash & 4.04 & 4.13 & 3.00 & 1.01 & 4.19 & \textbf{4.86} \\
 & DSpark & 4.63 & 4.90 & 3.38 & 1.02 & 5.05 & \textbf{5.26} \\
\bottomrule
\end{tabular}
\end{table}
\section{Conclusion}
\label{sec:conclusion}

We introduced Block Verification-aware loss (BV loss), a sequence-level drafter training objective derived from the block verification acceptance rule. The objective directly targets expected accepted draft length and recovers TV loss in the single-token case. A reformulation on target-generated trajectories enables training without differentiating through draft sampling or verification, while a blockwise logarithmic surrogate and early-training annealing support training from scratch.

Across math, code, and chat benchmarks, BV loss improves the mean number of tokens retained per verification call by 13.0--21.0\% over CE training for DFlash and DSpark with Qwen3-4B and Qwen3-8B under block
verification, with corresponding gains in decoding throughput. It also outperforms tokenwise acceptance objectives such as TV and LK, and its gains extend to token verification and greedy decoding without changing the inference procedure. Our analyses show improved retention beyond the first token and demonstrate that training token accuracy alone is an incomplete proxy for accepted draft length. Together, these findings highlight the benefit of training parallel drafters for sequence-level acceptance rather than optimizing individual token predictions in isolation.
\section{Limitations}
\label{sec:limitations}

Our study focuses on parallel block drafters, while the broader principle of verification-aligned training may apply beyond this setting. In particular, BV loss may also be applicable to autoregressive drafters such as EAGLE~\citep{eagle,li2024eagle2,li2025eagle3}, although our empirical evaluation is limited to diffusion-based drafters. More broadly, this perspective could be adapted to tree-based proposal and verification schemes, including methods based on Traversal Verification~\citep{weng2025traversal}. Exploring these extensions is beyond the scope of this work and is left for future research.

\subsection*{AI use statement}

In this work, we used generative AI tools to assist with writing experimental
code. The authors formulated the research ideas and hypotheses, designed
the experiments, and interpreted the results without generative AI assistance.
Additionally, we used generative AI tools to draft and revise the manuscript
and to prepare its \LaTeX{} source.
The authors verified the evaluation code and reviewed the final manuscript,
including the AI-assisted text. We take responsibility for the final content
of this work, including all text, claims, and artifacts produced with the
aid of generative AI.

\subsection*{Reproducibility statement}

Section~\ref{sec:method} defines the BV objective and its training
formulation, with the assumptions and supporting derivations detailed
in Appendix~\ref{sec:appendix_method}.
Section~\ref{sec:experimental_setup} describes the models, benchmarks,
and evaluation metrics.
Appendix~\ref{sec:appendix_implementation} documents the compared objectives,
training data and preprocessing, optimization settings, annealing schedule,
and numerical implementation.
The diagnostic evaluation and speedup measurement protocols are detailed
in Appendices~\ref{sec:diagnostic_evaluations} and~\ref{sec:speedup_measurement},
respectively, and Appendix~\ref{sec:full_results} provides the full
per-benchmark results for the compared objectives.

\bibliography{iclr2027_conference}
\bibliographystyle{iclr2027_conference}

\newpage
\appendix
\section{Additional Implementation Details}
\label{sec:appendix_implementation}

\subsection{Training Objectives}
\label{sec:appendix_objectives}

We compare CE, KL, RKL, TV, and LK as tokenwise alternatives to BV.
The tokenwise losses follow Eqs.~\eqref{eq:token_losses}
and~\eqref{eq:lk_losses} and use the position weights in
Eq.~\eqref{eq:token_weighting}, with $\eta=7$ for DFlash and $\eta=4$
for DSpark. In training, the weighted losses are normalized by the
total weight over valid prediction positions. The target model remains
frozen for every objective.

\paragraph{CE, KL, and RKL.}
CE predicts the stored target-generated token at each supervised position.
KL instead matches the full target distribution by minimizing
$D_{\mathrm{KL}}(p_i\|q_i)$; with a frozen target, this is equivalent to
soft-target cross-entropy up to a constant. RKL reverses the direction
to $D_{\mathrm{KL}}(q_i\|p_i)$, weighting the discrepancy by the draft
distribution. Both divergences use target and draft probabilities aligned
to the same prediction position and are evaluated over the full vocabulary.

\paragraph{TV and LK.}
TV minimizes the total variation distance between target and draft
distributions, equivalently maximizing their tokenwise overlap
$\sum_v\min\{p_i(v),q_i(v)\}$. LK applies a negative logarithm to this
overlap, increasing the gradient scale when acceptance is low.
These objectives also use full-vocabulary distributions. 

\paragraph{BV loss.}
BV loss aggregates prefix-retention scores into an estimate of accepted draft
length, coupling predictions across positions. The main BV runs use
the vocabulary-integrated estimate in Appendix~\ref{sec:integration_details}
within a blockwise log objective. The initial aggregation schedule is
described in Appendix~\ref{sec:annealing_details}.

\subsection{Training and Evaluation Protocol}
\label{sec:training_protocol}

\paragraph{Training data and optimization.}
We build the training corpora from the same 100K conversations selected
from Open-PerfectBlend,
an open-source reproduction of the instruction dataset described in
PerfectBlend~\citep{xu2024perfect}.
For each target (Qwen3-4B or Qwen3-8B), we regenerate the assistant
responses at $T=1$. The main-table runs for a given target share this
corpus and train from scratch for six epochs. They use AdamW with peak learning
rate $6\times10^{-4}$, global batch size 32, local batch size four,
4\% learning-rate warmup followed by cosine decay, and seed 42.
The maximum training sequence length is 3,072 tokens.

\paragraph{Drafter configurations.}
All compared models use five-layer drafters with $B=15$ draft tokens, where the anchor token is not counted toward $B$.
DSpark retains its causal Markov correction, and the losses are applied to the resulting corrected
proposal distribution. The correction parameters are trained with the
drafter, while the confidence head is disabled across the compared
DSpark objectives.

\paragraph{Decoding and reporting.}
All objectives are evaluated with a 256-token generation limit and
thinking disabled, under greedy decoding or full-vocabulary sampling
at $T=1$. For generation call $j$, let
$R_j$ be the number of retained tokens, including correction/bonus tokens
when present, and let $V_j$ be its number of verification calls. For
benchmark $d$, the table metric is
\[
 \tau_d^{\mathrm{table}}=\frac{\sum_{j\in d}R_j}{\sum_{j\in d}V_j},
 \qquad
 \mathrm{Avg.}=\frac{1}{7}\sum_{d=1}^{7}\tau_d^{\mathrm{table}}.
\]
We pool verification calls within each benchmark and give equal weight
to the seven benchmarks in the final average. Tables abbreviate
$\tau_d^{\mathrm{table}}$ as $\tau$.
Counts exclude the initial prefill token and are recorded before the
final generation is cropped to the output budget.
The evaluation includes 512 GSM8K, 500 MATH-500, 30 AIME25, 164 HumanEval,
257 MBPP, 256 LiveCodeBench, and 80 two-turn MT-Bench examples,
giving 1,879 generation calls per decoding condition.
Appendix~\ref{sec:speedup_measurement} describes the throughput measurement.

\subsection{Initial Annealing Schedule}
\label{sec:annealing_details}

The main BV runs linearly increase $\beta$ from zero to one over the
first three epochs, then use the blockwise log loss for the remaining
three epochs. This schedule applies Eqs.~\eqref{eq:annealing}
and~\eqref{eq:annealing_gradient} to the vocabulary-integrated scores
$\bar{\alpha}_i$ in place of $\widetilde{\alpha}_i$ for both drafters.
It gradually shifts from equal weighting of prefix log-scores to
weighting proportional to their acceptance scores. This aggregation
schedule is separate from the learning-rate warmup; unannealed BV runs
use $\beta=1$ throughout.

During the ramp, the implementation differentiates a negative weighted
mean of log-scores while holding the normalized power weights fixed in
the backward pass. For positive scores with inactive numerical floors,
this yields the gradient in Eq.~\eqref{eq:annealing_gradient}, although
the logged scalar need not equal Eq.~\eqref{eq:annealing}.
DFlash floors scores at $10^{-6}$ during the ramp; DSpark computes the
annealed log-scores without this floor.

\subsection{Numerical Stability}
\label{sec:numerical_evaluation}

BV cumulative ratios and prefix minima are computed in FP32 log space,
with gradients propagated through the prefix states. In the block-log
phase, log summands are floored at $-80$ and the summed length estimate
at $10^{-6}$. The theoretical identities use the unclamped expressions.

\subsection{Diagnostic Evaluations}
\label{sec:diagnostic_evaluations}

The analyses in Figures~\ref{fig:position_acceptance}
and~\ref{fig:training_dynamics} compare DFlash (CE) and DFlash + Ours (BV) to examine how the
training objective affects retained prefixes and token predictions.

\paragraph{Prefix-retention analysis (Figure~\ref{fig:position_acceptance}).}
The final six-epoch DFlash CE and annealed BV checkpoints are evaluated
on 512 GSM8K prompts with Qwen3-4B, block verification, $T=1$, and $B=15$.
Accepted-draft counts exclude correction/bonus tokens and prefill,
respect EOS, and precede final output-budget cropping.
We pool verification calls across prompts to estimate unconditional
prefix survival, $\Pr_{\mathrm{BV}}(\tau\ge i)$, in the left panel.
The right panel reports the ratio to
$\Pr_{\mathrm{BV}}(\tau\ge i-1)$, with denominator one at $i=1$,
giving the probability of extending a retained prefix.
These statistics average over each checkpoint's decoding trajectories;
unlike $\alpha_i(y_{1:i})$ in Eq.~\eqref{eq:block_acceptance}, they
do not fix the proposed token values.

\paragraph{Training dynamics (Figure~\ref{fig:training_dynamics}).}
Both models are evaluated at 4,096 fixed contexts across all six epochs.
We construct these contexts by uniformly sampling 512 of the 98,499
retained Qwen3-4B training examples (seed 0) and choosing eight full
assistant blocks per example. Cached labels and the 3,072-token length
limit are preserved. Blocks may overlap, but each prefix is evaluated
independently and the same blocks are used for both models.
Token accuracy pools 61,440 unweighted argmax predictions against the
stored target tokens. Accepted draft length averages one block-verification
draw per context using full-vocabulary target and draft distributions
($T=1$, $B=15$), respecting EOS and excluding anchors and correction/bonus
tokens. The CE curve uses the original epoch checkpoints; the BV curve
uses a same-seed repeat with checkpoints retained at every epoch.
The repeat preserves global batch size 32 and the optimizer schedule
across a change in GPU count. Both curves use a single training seed.

\subsection{Speedup Measurement}
\label{sec:speedup_measurement}

\paragraph{Execution and timing.}
Each inference run uses a single NVIDIA B200 GPU at batch size one,
with BF16 model weights and FlashAttention-2. Both speculative decoding
(SD) and target-only autoregressive decoding (AR) run in inference mode
with fresh KV caches, matching prompts, sampling and stopping rules,
and a 256-token generation limit.
We time the same decoding implementation used to measure $\tau$, with
CUDA synchronization at the boundaries of each timed region.
Elapsed time includes prefill, draft generation,
verification, sampling, and cache updates; model loading, prompt formatting,
tokenization, and file I/O are excluded. After a warmup for each dataset, we make one measured pass over its evaluation examples.

\paragraph{Aggregation.}
For generation call $i$ in benchmark $d$, let $n_i^m$ and $t_i^m$ denote
the generated token count and total elapsed time for
$m\in\{\mathrm{SD},\mathrm{AR}\}$. We pool token counts and times within
each benchmark before taking the throughput ratio:
\[
 \mathrm{Speedup}_d =
 \frac{\sum_{i\in d} n_i^{\mathrm{SD}} / \sum_{i\in d} t_i^{\mathrm{SD}}}
      {\sum_{i\in d} n_i^{\mathrm{AR}} / \sum_{i\in d} t_i^{\mathrm{AR}}}.
\]

\section{Additional Derivations}
\label{sec:appendix_method}

We first establish the target-path identity underlying BV and illustrate
why tokenwise overlap does not determine block acceptance. The remaining
derivations explain prefix credit, vocabulary integration, and the
relationship between the population objective and its training surrogate.

\subsection{Expected Acceptance Length on Target-Generated Paths}
\label{sec:estimator_details}

We follow Section~\ref{sec:bv_identity}: $y$ denotes a proposed draft
and $y^\prime$ a target-generated training path. Fix a verification
context $c$, a frozen target, and positive target and proposal probabilities.
Let $q_i$ be the conditionals of the actual proposal process and $p_i$
the target conditionals used to generate the training paths.
For any prefix $x_{1:i}$, write
$p(x_{1:i}\mid c)=\prod_{j=1}^{i}p_j(x_j)$ and
$q(x_{1:i}\mid c)=\prod_{j=1}^{i}q_j(x_j)$.

The block-verification rule in Eq.~\eqref{eq:block_acceptance}
gives the retention probability conditional on the proposed prefix,
averaging over proposal suffixes and verifier randomness~\citep{sun2025block}.
Here, conditioning fixes the proposed token values rather than an
earlier acceptance event. Multiplying by the cumulative ratio in
Eq.~\eqref{eq:prefix_states} gives
$r_i\alpha_i=\min\{r_i,r_{i-1}\alpha_{i-1}\}
=\widetilde{\alpha}_i$.
Consequently, the joint mass of proposing and retaining a prefix is
\begin{equation}
 \begin{aligned}
 \Pr_{\mathrm{BV}}(y_{1:i}=x_{1:i},\tau\ge i\mid c)
 &=q(x_{1:i}\mid c)\alpha_i(x_{1:i})\\
 &=p(x_{1:i}\mid c)\widetilde{\alpha}_i(x_{1:i}).
 \end{aligned}
 \label{eq:prefix_mass}
\end{equation}
Summing over prefixes yields
$\Pr_{\mathrm{BV}}(\tau\ge i\mid c)
=\mathbb{E}_{y^\prime\sim p}[\widetilde{\alpha}_i\mid c]$.
Applying the tail-sum identity establishes
Eq.~\eqref{eq:target_prefix_identity}, and normalizing by $B$ gives
the training form in Eq.~\eqref{eq:linear_bv_train}.
The same result holds after averaging over a fixed distribution of
training contexts. Under these sampling conditions, a minibatch average
of $1-B^{-1}\sum_i\widetilde{\alpha}_i$ is an unbiased estimate of the
population BV loss.

For $B=1$, the expected score is
$\sum_v p_1(v)\min\{1,q_1(v)/p_1(v)\}
=\sum_v\min\{p_1(v),q_1(v)\}$.
Thus the linear BV loss recovers TV exactly, while the fully integrated
block-log loss recovers LK.

\subsection{Equal Tokenwise Overlap, Different Block Acceptance}
\label{sec:token_block_gap}

Consider a two-token block with independent binary target distributions,
$p_1(a)=0.2$ and $p_2(a)=0.1$. The two proposals below have identical
tokenwise overlaps; in each distribution, the other token receives the
remaining probability. Enumerating the four possible blocks using
Eq.~\eqref{eq:prefix_mass} gives
\begin{center}
\begin{tabular}{lcc}
\toprule
Quantity & Draft A & Draft B\\
\midrule
$(q_1(a),q_2(a))$ & $(0.3,0.2)$ & $(0.1,0.2)$\\
Tokenwise overlaps & $(0.9,0.9)$ & $(0.9,0.9)$\\
BV prefix survival & $(0.9,0.83)$ & $(0.9,0.89)$\\
BV $\mathbb{E}[\tau]$ & $1.73$ & $1.79$\\
\bottomrule
\end{tabular}
\end{center}
Both proposals have TV loss $0.1$ and LK loss $-\log(0.9)$ at each
position, yet their expected block acceptance lengths differ. Under
token verification, both have survival probabilities $(0.9,0.81)$ and
expected accepted draft length $1.71$. Tokenwise overlap therefore
determines token-verification acceptance in this example, but not block
acceptance. The latter depends on how likelihood ratios combine across
prefixes, which a fixed weighted sum of these tokenwise losses cannot
distinguish.

\subsection{Prefix Credit and Conditional-Logit Gradients}
\label{sec:prefix_gradient_details}

For a fixed target-generated block, let
$S=\sum_i\widetilde{\alpha}_i>0$ and let $n_i$ be the active prefix
minimum from Section~\ref{sec:gradients}. Away from ties,
Eq.~\eqref{eq:acceptance_gradient} gives
\begin{equation}
 \nabla(\log B-\log S)
 =-\sum_{j=1}^{B}c_j\nabla\log q_j(y^\prime_j),
 \qquad
 c_j=\frac{\sum_{i:n_i\ge j}\widetilde{\alpha}_i}{S}\in[0,1].
 \label{eq:prefix_credit}
\end{equation}
The coefficient $c_j$ collects the scores whose active prefix minimum
depends on position $j$. For conditional softmax logits $z_{j,v}$,
holding the path and other conditional logits fixed,
\begin{equation}
 \frac{\partial(\log B-\log S)}{\partial z_{j,v}}
 =-c_j\bigl(\mathbf{1}\{v=y^\prime_j\}-q_j(v)\bigr),
 \qquad
 \left|\frac{\partial(\log B-\log S)}{\partial z_{j,v}}\right|\le1.
 \label{eq:conditional_logit_bound}
\end{equation}
For this sampled-label surrogate, log scaling removes the overall
small-score factor while keeping the conditional-logit derivatives
bounded. At ties, convex
combinations of the active gradients satisfy the same bound.
This statement concerns conditional logits; it does not bound gradients
with respect to shared neural-network parameters.

\subsection{Vocabulary-Integrated Length Estimation}
\label{sec:integration_details}

The sampled-label estimate is
$\widehat\tau_{\mathrm{sample}}=\sum_i\widetilde{\alpha}_i(y^\prime_{1:i})$.
To use the full target distribution at the current position, we can
integrate out that token while keeping the sampled prefix fixed:
\begin{equation}
 \begin{aligned}
 \bar{\alpha}_i
 &:=\mathbb{E}_{y^\prime_i\sim p_i}
       [\widetilde{\alpha}_i\mid y^\prime_{<i},c]\\
 &=\sum_v\min\{\widetilde{\alpha}_{i-1}p_i(v),r_{i-1}q_i(v)\},
 \qquad \widehat\tau_{\mathrm{int}}=\sum_i\bar{\alpha}_i.
 \end{aligned}
 \label{eq:integrated_score}
\end{equation}
The formula follows by substituting
$r_i=r_{i-1}q_i(y^\prime_i)/p_i(y^\prime_i)$ into
Eq.~\eqref{eq:acceptance_score}. By total expectation, both length
estimates are unbiased for $\mathbb{E}_{\mathrm{BV}}[\tau\mid c]$
under the conditions of Appendix~\ref{sec:estimator_details}.
Only the current score is integrated: $r_i$ and
$\widetilde{\alpha}_i$ continue to evolve along the sampled target path,
preserving its prefix dependencies. The main BV runs use
$\widehat\tau_{\mathrm{int}}$ in both the blockwise log objective and
its annealed form.

\subsection{Population and Blockwise Log Objectives}
\label{sec:log_details}

The population objective in Eq.~\eqref{eq:population_bv_loss} takes
the negative logarithm after averaging accepted length. Practical
training instead averages the normalized negative log of each block estimate,
$\ell_{\mathrm{BV}}^{\log}=\log B-\log\widehat\tau$.
For a positive, unbiased length estimate without numerical floors,
Jensen's inequality gives
\begin{equation}
 \mathbb{E}[\ell_{\mathrm{BV}}^{\log}]
 \ge\log B-\log\mathbb{E}[\widehat\tau]
 =\mathcal L_{\mathrm{BV}}^{\log}.
 \label{eq:jensen}
\end{equation}
Thus, the expected blockwise log loss upper-bounds the population log
objective. An unbiased length estimate need not yield an unbiased log
loss or the same optimum within a restricted drafter family.
Both objectives share the ideal solution $q=p$, when it is
representable: every prefix score is one and $\widehat\tau=B$.
The annealing schedule returns to the blockwise log surrogate at
$\beta=1$.

\section{Additional Experimental Results}
\label{sec:appendix_experiments}

\subsection{Comparison of BV Training Recipes}
\label{sec:annealing_ablation}

Table~\ref{tab:annealing} compares CE, BV without annealing, and the
main annealed BV recipe. Importantly, BV already provides substantial
acceptance gains without annealing, showing that the annealing schedule is not essential for successful optimization. Annealing provides an additional improvement, yielding the best results across the reported settings.

On Qwen3-4B at $T=1$, unannealed BV increases
DFlash's mean $\tau$ from 4.09 to 4.74 (\textbf{+15.9\%}) and
DSpark's from 4.77 to 5.17 (\textbf{+8.4\%}), both under block
verification. Annealing further improves these values to 4.95 and 5.39,
respectively. Qwen3-8B shows the same pattern: unannealed BV reaches
4.80 for DFlash and 5.13 for DSpark, while the annealed recipes reach
4.98 and 5.43. A similar trend is observed under greedy decoding.
Overall, these results suggest that annealing is a useful optimization
aid rather than a prerequisite for BV training, consistent with the
discussion in Section~\ref{sec:log_loss} on increasing the relative
training weight of low-scoring prefixes early in training.

\begin{table}[!htbp]
\centering
\caption{BV and annealing ablation. Entries are average $\tau$ over the
seven benchmarks, with the counting convention of Table~\ref{tab:main_results}.
At $T=1$, both drafters use block verification.}
\label{tab:annealing}
\small
\setlength{\tabcolsep}{4pt}
\renewcommand{\arraystretch}{1.12}
\begin{tabular}{@{}l*{4}{c}@{\hspace{12pt}}*{4}{c}@{}}
\toprule
 & \multicolumn{4}{c}{Qwen3-4B} & \multicolumn{4}{c}{Qwen3-8B}\\
\cmidrule(lr){2-5}\cmidrule(lr){6-9}
 & \multicolumn{2}{c}{DFlash} & \multicolumn{2}{c}{DSpark} & \multicolumn{2}{c}{DFlash} & \multicolumn{2}{c}{DSpark}\\
\cmidrule(lr){2-3}\cmidrule(lr){4-5}\cmidrule(lr){6-7}\cmidrule(lr){8-9}
Training loss & $T=0$ & $T=1$ & $T=0$ & $T=1$ & $T=0$ & $T=1$ & $T=0$ & $T=1$\\
\midrule
Baseline (CE) & 4.65 & 4.09 & 5.21 & 4.77 & 4.64 & 4.15 & 5.10 & 4.71\\
BV w/o annealing & 4.98 & 4.74 & 5.37 & 5.17 & 5.07 & 4.80 & 5.36 & 5.13\\
BV with annealing (Ours) & \textbf{5.22} & \textbf{4.95} & \textbf{5.62} & \textbf{5.39} & \textbf{5.27} & \textbf{4.98} & \textbf{5.67} & \textbf{5.43}\\
\bottomrule
\end{tabular}
\end{table}

\subsection{Full Benchmark Results}
\label{sec:full_results}

Tables~\ref{tab:full_results} and~\ref{tab:full_results_qwen3_8b} provide
the per-benchmark results for every training objective with Qwen3-4B
and Qwen3-8B. They expand the averages in
Tables~\ref{tab:main_results} and~\ref{tab:token_verification} across
greedy decoding, token verification at $T=1$, and block verification at
$T=1$, using the same $\tau$ convention and autoregressive speedup
reference as Section~\ref{sec:experimental_setup}.
BV improves over CE in all 84 benchmark-level acceptance-length
comparisons across the two targets, two drafters, and three settings.
BV also achieves the highest mean acceptance length among the compared
objectives in each configuration. The gains over CE therefore extend
across individual math, code, and chat benchmarks, as well as their
aggregate results.

\begin{table}[!htb]
\centering
\caption{Full benchmark results for DFlash and DSpark on Qwen3-4B.
Here, $\tau$ is the reported verification-call length, including
correction/bonus when present; the Method uses draft-only $\tau$.
Avg.\ retains the reported seven-benchmark mean. Speedup is relative to
autoregressive decoding.
Bold marks the best value for each metric within
each drafter and setting..}
\label{tab:full_results}
\scriptsize
\setlength{\tabcolsep}{1.20pt}
\renewcommand{\arraystretch}{1.1}
\resizebox{\textwidth}{!}{%
\begin{tabular}{@{}l@{\hspace{4pt}}l@{\hspace{3pt}}*{16}{c}@{}}
\toprule
\multirow{2}{*}{Drafter} & \multirow{2}{*}{Loss}
 & \multicolumn{6}{c}{Math} & \multicolumn{6}{c}{Code}
 & \multicolumn{2}{c}{Chat} & \multicolumn{2}{c}{}\\
\cmidrule(lr){3-8}\cmidrule(lr){9-14}\cmidrule(lr){15-16}
 & & \multicolumn{2}{c}{GSM8K} & \multicolumn{2}{c}{MATH-500}
 & \multicolumn{2}{c}{AIME25} & \multicolumn{2}{c}{HumanEval}
 & \multicolumn{2}{c}{MBPP} & \multicolumn{2}{c}{LCB}
 & \multicolumn{2}{c}{MT-Bench} & \multicolumn{2}{c}{Avg.}\\
\midrule
\multicolumn{2}{@{}l@{\hspace{5pt}}}{$T=0$} & {\fontsize{6}{7}\selectfont Speedup} & $\tau$ & {\fontsize{6}{7}\selectfont Speedup} & $\tau$ & {\fontsize{6}{7}\selectfont Speedup} & $\tau$ & {\fontsize{6}{7}\selectfont Speedup} & $\tau$ & {\fontsize{6}{7}\selectfont Speedup} & $\tau$ & {\fontsize{6}{7}\selectfont Speedup} & $\tau$ & {\fontsize{6}{7}\selectfont Speedup} & $\tau$ & {\fontsize{6}{7}\selectfont Speedup} & $\tau$ \\
\midrule
\multirow{6}{*}{DFlash} & CE & 4.99\ensuremath{\times} & 6.32 & 4.46\ensuremath{\times} & 5.64 & 3.63\ensuremath{\times} & 4.50 & 3.53\ensuremath{\times} & 4.45 & 3.41\ensuremath{\times} & 4.27 & 3.59\ensuremath{\times} & 4.44 & 2.34\ensuremath{\times} & 2.94 & 3.71\ensuremath{\times} & 4.65 \\
 & KL & 5.05\ensuremath{\times} & 6.51 & 4.54\ensuremath{\times} & 5.83 & 3.68\ensuremath{\times} & 4.67 & 3.60\ensuremath{\times} & 4.61 & 3.46\ensuremath{\times} & 4.40 & 3.61\ensuremath{\times} & 4.56 & 2.40\ensuremath{\times} & 3.03 & 3.76\ensuremath{\times} & 4.80 \\
 & RKL & 4.52\ensuremath{\times} & 5.69 & 3.84\ensuremath{\times} & 4.81 & 3.09\ensuremath{\times} & 3.85 & 3.05\ensuremath{\times} & 3.82 & 2.99\ensuremath{\times} & 3.71 & 3.22\ensuremath{\times} & 3.98 & 2.00\ensuremath{\times} & 2.51 & 3.24\ensuremath{\times} & 4.05 \\
 & TV & 1.95\ensuremath{\times} & 2.42 & 2.01\ensuremath{\times} & 2.48 & 1.78\ensuremath{\times} & 2.19 & 1.39\ensuremath{\times} & 1.71 & 1.44\ensuremath{\times} & 1.77 & 1.34\ensuremath{\times} & 1.63 & 1.32\ensuremath{\times} & 1.64 & 1.60\ensuremath{\times} & 1.98 \\
 & LK & 5.08\ensuremath{\times} & 6.55 & 4.52\ensuremath{\times} & 5.81 & 3.64\ensuremath{\times} & 4.63 & 3.59\ensuremath{\times} & 4.61 & 3.45\ensuremath{\times} & 4.39 & 3.61\ensuremath{\times} & 4.55 & 2.38\ensuremath{\times} & 2.99 & 3.75\ensuremath{\times} & 4.79 \\
 & \textbf{BV (Ours)} & \textbf{5.57\ensuremath{\times}} & \textbf{7.34} & \textbf{4.84\ensuremath{\times}} & \textbf{6.36} & \textbf{3.94\ensuremath{\times}} & \textbf{5.15} & \textbf{3.79\ensuremath{\times}} & \textbf{4.93} & \textbf{3.61\ensuremath{\times}} & \textbf{4.70} & \textbf{3.76\ensuremath{\times}} & \textbf{4.87} & \textbf{2.54\ensuremath{\times}} & \textbf{3.20} & \textbf{4.01\ensuremath{\times}} & \textbf{5.22} \\
\cmidrule(lr){1-18}
\multirow{6}{*}{DSpark} & CE & 5.28\ensuremath{\times} & 7.03 & 4.84\ensuremath{\times} & 6.38 & 3.87\ensuremath{\times} & 5.08 & 3.84\ensuremath{\times} & 5.07 & 3.66\ensuremath{\times} & 4.79 & 3.78\ensuremath{\times} & 4.92 & 2.50\ensuremath{\times} & 3.23 & 3.97\ensuremath{\times} & 5.21 \\
 & KL & 5.55\ensuremath{\times} & 7.42 & 5.07\ensuremath{\times} & 6.77 & 4.08\ensuremath{\times} & 5.41 & 4.00\ensuremath{\times} & 5.30 & 3.81\ensuremath{\times} & 5.03 & 3.88\ensuremath{\times} & 5.11 & 2.63\ensuremath{\times} & 3.39 & 4.15\ensuremath{\times} & 5.49 \\
 & RKL & 5.19\ensuremath{\times} & 6.84 & 4.50\ensuremath{\times} & 5.90 & 3.66\ensuremath{\times} & 4.79 & 3.51\ensuremath{\times} & 4.61 & 3.41\ensuremath{\times} & 4.44 & 3.53\ensuremath{\times} & 4.56 & 2.27\ensuremath{\times} & 2.93 & 3.73\ensuremath{\times} & 4.87 \\
 & TV & 1.95\ensuremath{\times} & 2.52 & 2.16\ensuremath{\times} & 2.78 & 1.93\ensuremath{\times} & 2.48 & 1.58\ensuremath{\times} & 2.03 & 1.58\ensuremath{\times} & 2.04 & 1.53\ensuremath{\times} & 1.95 & 1.43\ensuremath{\times} & 1.82 & 1.74\ensuremath{\times} & 2.23 \\
 & LK & \textbf{5.71\ensuremath{\times}} & 7.53 & 5.17\ensuremath{\times} & 6.79 & 4.21\ensuremath{\times} & 5.49 & \textbf{4.08\ensuremath{\times}} & \textbf{5.35} & \textbf{3.90\ensuremath{\times}} & 5.04 & 3.99\ensuremath{\times} & 5.14 & 2.63\ensuremath{\times} & \textbf{3.41} & 4.24\ensuremath{\times} & 5.54 \\
 & \textbf{BV (Ours)} & \textbf{5.71\ensuremath{\times}} & \textbf{7.75} & \textbf{5.25\ensuremath{\times}} & \textbf{6.91} & \textbf{4.33\ensuremath{\times}} & \textbf{5.64} & \textbf{4.08\ensuremath{\times}} & 5.33 & \textbf{3.90\ensuremath{\times}} & \textbf{5.07} & \textbf{4.03\ensuremath{\times}} & \textbf{5.20} & \textbf{2.64\ensuremath{\times}} & \textbf{3.41} & \textbf{4.28\ensuremath{\times}} & \textbf{5.62} \\
\midrule
\multicolumn{2}{@{}l@{\hspace{5pt}}}{$T=1$, token verify} & {\fontsize{6}{7}\selectfont Speedup} & $\tau$ & {\fontsize{6}{7}\selectfont Speedup} & $\tau$ & {\fontsize{6}{7}\selectfont Speedup} & $\tau$ & {\fontsize{6}{7}\selectfont Speedup} & $\tau$ & {\fontsize{6}{7}\selectfont Speedup} & $\tau$ & {\fontsize{6}{7}\selectfont Speedup} & $\tau$ & {\fontsize{6}{7}\selectfont Speedup} & $\tau$ & {\fontsize{6}{7}\selectfont Speedup} & $\tau$ \\
\midrule
\multirow{6}{*}{DFlash} & CE & 4.21\ensuremath{\times} & 5.40 & 3.70\ensuremath{\times} & 4.74 & 2.91\ensuremath{\times} & 3.70 & 3.09\ensuremath{\times} & 3.97 & 2.93\ensuremath{\times} & 3.71 & 3.15\ensuremath{\times} & 3.95 & 2.06\ensuremath{\times} & 2.59 & 3.15\ensuremath{\times} & 4.01 \\
 & KL & 4.29\ensuremath{\times} & 5.53 & 3.79\ensuremath{\times} & 4.80 & 2.97\ensuremath{\times} & 3.75 & 3.27\ensuremath{\times} & 4.19 & 3.04\ensuremath{\times} & 3.84 & 3.23\ensuremath{\times} & 4.02 & 2.08\ensuremath{\times} & 2.62 & 3.24\ensuremath{\times} & 4.11 \\
 & RKL & 4.15\ensuremath{\times} & 5.31 & 3.53\ensuremath{\times} & 4.47 & 2.88\ensuremath{\times} & 3.60 & 2.96\ensuremath{\times} & 3.73 & 2.89\ensuremath{\times} & 3.62 & 3.11\ensuremath{\times} & 3.85 & 1.93\ensuremath{\times} & 2.43 & 3.06\ensuremath{\times} & 3.86 \\
 & TV & 1.86\ensuremath{\times} & 2.37 & 1.92\ensuremath{\times} & 2.44 & 1.71\ensuremath{\times} & 2.17 & 1.35\ensuremath{\times} & 1.72 & 1.39\ensuremath{\times} & 1.76 & 1.30\ensuremath{\times} & 1.62 & 1.31\ensuremath{\times} & 1.63 & 1.55\ensuremath{\times} & 1.96 \\
 & LK & 4.41\ensuremath{\times} & 5.61 & 3.82\ensuremath{\times} & 4.85 & 3.05\ensuremath{\times} & 3.84 & 3.33\ensuremath{\times} & 4.24 & 3.07\ensuremath{\times} & 3.85 & 3.33\ensuremath{\times} & 4.15 & 2.14\ensuremath{\times} & 2.69 & 3.31\ensuremath{\times} & 4.17 \\
 & \textbf{BV (Ours)} & \textbf{5.14\ensuremath{\times}} & \textbf{6.70} & \textbf{4.44\ensuremath{\times}} & \textbf{5.74} & \textbf{3.59\ensuremath{\times}} & \textbf{4.61} & \textbf{3.66\ensuremath{\times}} & \textbf{4.69} & \textbf{3.47\ensuremath{\times}} & \textbf{4.49} & \textbf{3.63\ensuremath{\times}} & \textbf{4.67} & \textbf{2.45\ensuremath{\times}} & \textbf{3.09} & \textbf{3.77\ensuremath{\times}} & \textbf{4.85} \\
\cmidrule(lr){1-18}
\multirow{6}{*}{DSpark} & CE & 4.49\ensuremath{\times} & 6.25 & 4.06\ensuremath{\times} & 5.63 & 3.27\ensuremath{\times} & 4.49 & 3.30\ensuremath{\times} & 4.57 & 3.15\ensuremath{\times} & 4.31 & 3.20\ensuremath{\times} & 4.31 & 2.18\ensuremath{\times} & 2.94 & 3.38\ensuremath{\times} & 4.64 \\
 & KL & 4.71\ensuremath{\times} & 6.52 & 4.29\ensuremath{\times} & 5.91 & 3.45\ensuremath{\times} & 4.71 & 3.49\ensuremath{\times} & 4.79 & 3.27\ensuremath{\times} & 4.45 & 3.36\ensuremath{\times} & 4.53 & 2.21\ensuremath{\times} & 3.00 & 3.54\ensuremath{\times} & 4.84 \\
 & RKL & 4.62\ensuremath{\times} & 6.38 & 3.97\ensuremath{\times} & 5.48 & 3.24\ensuremath{\times} & 4.48 & 3.26\ensuremath{\times} & 4.50 & 3.18\ensuremath{\times} & 4.33 & 3.32\ensuremath{\times} & 4.48 & 2.10\ensuremath{\times} & 2.84 & 3.38\ensuremath{\times} & 4.64 \\
 & TV & 1.81\ensuremath{\times} & 2.50 & 1.99\ensuremath{\times} & 2.73 & 1.74\ensuremath{\times} & 2.39 & 1.48\ensuremath{\times} & 2.03 & 1.49\ensuremath{\times} & 2.02 & 1.45\ensuremath{\times} & 1.94 & 1.34\ensuremath{\times} & 1.80 & 1.62\ensuremath{\times} & 2.20 \\
 & LK & 4.97\ensuremath{\times} & 6.74 & 4.38\ensuremath{\times} & 5.99 & \textbf{3.57\ensuremath{\times}} & \textbf{4.85} & 3.59\ensuremath{\times} & 4.88 & 3.41\ensuremath{\times} & 4.58 & 3.55\ensuremath{\times} & 4.70 & 2.30\ensuremath{\times} & 3.11 & 3.68\ensuremath{\times} & 4.98 \\
 & \textbf{BV (Ours)} & \textbf{5.14\ensuremath{\times}} & \textbf{7.15} & \textbf{4.60\ensuremath{\times}} & \textbf{6.33} & 3.53\ensuremath{\times} & \textbf{4.85} & \textbf{3.69\ensuremath{\times}} & \textbf{5.08} & \textbf{3.54\ensuremath{\times}} & \textbf{4.85} & \textbf{3.57\ensuremath{\times}} & \textbf{4.88} & \textbf{2.38\ensuremath{\times}} & \textbf{3.23} & \textbf{3.78\ensuremath{\times}} & \textbf{5.20} \\
\midrule
\multicolumn{2}{@{}l@{\hspace{5pt}}}{$T=1$, block verify} & {\fontsize{6}{7}\selectfont Speedup} & $\tau$ & {\fontsize{6}{7}\selectfont Speedup} & $\tau$ & {\fontsize{6}{7}\selectfont Speedup} & $\tau$ & {\fontsize{6}{7}\selectfont Speedup} & $\tau$ & {\fontsize{6}{7}\selectfont Speedup} & $\tau$ & {\fontsize{6}{7}\selectfont Speedup} & $\tau$ & {\fontsize{6}{7}\selectfont Speedup} & $\tau$ & {\fontsize{6}{7}\selectfont Speedup} & $\tau$ \\
\midrule
\multirow{6}{*}{DFlash} & CE & 4.32\ensuremath{\times} & 5.55 & 3.78\ensuremath{\times} & 4.82 & 3.05\ensuremath{\times} & 3.88 & 3.14\ensuremath{\times} & 4.01 & 2.97\ensuremath{\times} & 3.76 & 3.15\ensuremath{\times} & 3.96 & 2.09\ensuremath{\times} & 2.65 & 3.21\ensuremath{\times} & 4.09 \\
 & KL & 4.32\ensuremath{\times} & 5.63 & 3.79\ensuremath{\times} & 4.91 & 3.05\ensuremath{\times} & 3.92 & 3.22\ensuremath{\times} & 4.20 & 3.01\ensuremath{\times} & 3.87 & 3.15\ensuremath{\times} & 4.00 & 2.10\ensuremath{\times} & 2.67 & 3.23\ensuremath{\times} & 4.17 \\
 & RKL & 4.22\ensuremath{\times} & 5.35 & 3.57\ensuremath{\times} & 4.50 & 2.82\ensuremath{\times} & 3.49 & 2.99\ensuremath{\times} & 3.77 & 2.92\ensuremath{\times} & 3.64 & 3.16\ensuremath{\times} & 3.91 & 1.94\ensuremath{\times} & 2.46 & 3.09\ensuremath{\times} & 3.87 \\
 & TV & 1.91\ensuremath{\times} & 2.39 & 1.95\ensuremath{\times} & 2.43 & 1.74\ensuremath{\times} & 2.16 & 1.37\ensuremath{\times} & 1.71 & 1.42\ensuremath{\times} & 1.76 & 1.33\ensuremath{\times} & 1.62 & 1.30\ensuremath{\times} & 1.62 & 1.57\ensuremath{\times} & 1.96 \\
 & LK & 4.48\ensuremath{\times} & 5.74 & 3.94\ensuremath{\times} & 5.04 & 3.08\ensuremath{\times} & 3.90 & 3.29\ensuremath{\times} & 4.24 & 3.12\ensuremath{\times} & 3.93 & 3.31\ensuremath{\times} & 4.14 & 2.16\ensuremath{\times} & 2.73 & 3.34\ensuremath{\times} & 4.25 \\
 & \textbf{BV (Ours)} & \textbf{5.27\ensuremath{\times}} & \textbf{6.85} & \textbf{4.56\ensuremath{\times}} & \textbf{5.92} & \textbf{3.78\ensuremath{\times}} & \textbf{4.80} & \textbf{3.70\ensuremath{\times}} & \textbf{4.79} & \textbf{3.52\ensuremath{\times}} & \textbf{4.53} & \textbf{3.61\ensuremath{\times}} & \textbf{4.68} & \textbf{2.42\ensuremath{\times}} & \textbf{3.09} & \textbf{3.84\ensuremath{\times}} & \textbf{4.95} \\
\cmidrule(lr){1-18}
\multirow{6}{*}{DSpark} & CE & 4.66\ensuremath{\times} & 6.47 & 4.20\ensuremath{\times} & 5.80 & 3.39\ensuremath{\times} & 4.67 & 3.37\ensuremath{\times} & 4.65 & 3.21\ensuremath{\times} & 4.38 & 3.24\ensuremath{\times} & 4.40 & 2.20\ensuremath{\times} & 3.01 & 3.47\ensuremath{\times} & 4.77 \\
 & KL & 4.99\ensuremath{\times} & 6.80 & 4.49\ensuremath{\times} & 6.08 & 3.50\ensuremath{\times} & 4.77 & 3.62\ensuremath{\times} & 4.91 & 3.40\ensuremath{\times} & 4.56 & 3.45\ensuremath{\times} & 4.57 & 2.26\ensuremath{\times} & 3.10 & 3.67\ensuremath{\times} & 4.97 \\
 & RKL & 4.76\ensuremath{\times} & 6.51 & 4.08\ensuremath{\times} & 5.57 & 3.23\ensuremath{\times} & 4.37 & 3.31\ensuremath{\times} & 4.53 & 3.27\ensuremath{\times} & 4.41 & 3.37\ensuremath{\times} & 4.51 & 2.13\ensuremath{\times} & 2.90 & 3.45\ensuremath{\times} & 4.68 \\
 & TV & 1.88\ensuremath{\times} & 2.50 & 2.05\ensuremath{\times} & 2.74 & 1.82\ensuremath{\times} & 2.42 & 1.54\ensuremath{\times} & 2.05 & 1.52\ensuremath{\times} & 2.02 & 1.48\ensuremath{\times} & 1.95 & 1.34\ensuremath{\times} & 1.80 & 1.66\ensuremath{\times} & 2.21 \\
 & LK & 5.10\ensuremath{\times} & 7.00 & 4.50\ensuremath{\times} & 6.19 & 3.78\ensuremath{\times} & 5.22 & 3.67\ensuremath{\times} & 5.02 & 3.50\ensuremath{\times} & 4.73 & 3.53\ensuremath{\times} & 4.71 & 2.31\ensuremath{\times} & 3.16 & 3.77\ensuremath{\times} & 5.15 \\
 & \textbf{BV (Ours)} & \textbf{5.22\ensuremath{\times}} & \textbf{7.28} & \textbf{4.74\ensuremath{\times}} & \textbf{6.60} & \textbf{3.83\ensuremath{\times}} & \textbf{5.39} & \textbf{3.73\ensuremath{\times}} & \textbf{5.26} & \textbf{3.55\ensuremath{\times}} & \textbf{4.91} & \textbf{3.66\ensuremath{\times}} & \textbf{5.05} & \textbf{2.39\ensuremath{\times}} & \textbf{3.26} & \textbf{3.88\ensuremath{\times}} & \textbf{5.39} \\
\bottomrule
\end{tabular}
}
\end{table}

\begin{table}[!htb]
\centering
\caption{Full benchmark results for DFlash and DSpark on Qwen3-8B.
Here, $\tau$ is the reported verification-call length, including
correction/bonus when present; the Method uses draft-only $\tau$.
Avg.\ retains the reported seven-benchmark mean. Speedup is relative to
autoregressive decoding.
Bold marks the best value for each metric within
each drafter and setting.}
\label{tab:full_results_qwen3_8b}
\scriptsize
\setlength{\tabcolsep}{1.20pt}
\renewcommand{\arraystretch}{1.1}
\resizebox{\textwidth}{!}{%
\begin{tabular}{@{}l@{\hspace{4pt}}l@{\hspace{3pt}}*{16}{c}@{}}
\toprule
\multirow{2}{*}{Drafter} & \multirow{2}{*}{Loss}
 & \multicolumn{6}{c}{Math} & \multicolumn{6}{c}{Code}
 & \multicolumn{2}{c}{Chat} & \multicolumn{2}{c}{}\\
\cmidrule(lr){3-8}\cmidrule(lr){9-14}\cmidrule(lr){15-16}
 & & \multicolumn{2}{c}{GSM8K} & \multicolumn{2}{c}{MATH-500}
 & \multicolumn{2}{c}{AIME25} & \multicolumn{2}{c}{HumanEval}
 & \multicolumn{2}{c}{MBPP} & \multicolumn{2}{c}{LCB}
 & \multicolumn{2}{c}{MT-Bench} & \multicolumn{2}{c}{Avg.}\\
\midrule
\multicolumn{2}{@{}l@{\hspace{5pt}}}{$T=0$} & {\fontsize{6}{7}\selectfont Speedup} & $\tau$ & {\fontsize{6}{7}\selectfont Speedup} & $\tau$ & {\fontsize{6}{7}\selectfont Speedup} & $\tau$ & {\fontsize{6}{7}\selectfont Speedup} & $\tau$ & {\fontsize{6}{7}\selectfont Speedup} & $\tau$ & {\fontsize{6}{7}\selectfont Speedup} & $\tau$ & {\fontsize{6}{7}\selectfont Speedup} & $\tau$ & {\fontsize{6}{7}\selectfont Speedup} & $\tau$ \\
\midrule
\multirow{6}{*}{DFlash} & CE & 4.70\ensuremath{\times} & 6.11 & 4.28\ensuremath{\times} & 5.57 & 3.53\ensuremath{\times} & 4.55 & 3.54\ensuremath{\times} & 4.61 & 3.26\ensuremath{\times} & 4.20 & 3.58\ensuremath{\times} & 4.58 & 2.32\ensuremath{\times} & 2.88 & 3.60\ensuremath{\times} & 4.64 \\
 & KL & 4.86\ensuremath{\times} & 6.37 & 4.42\ensuremath{\times} & 5.79 & 3.59\ensuremath{\times} & 4.67 & 3.69\ensuremath{\times} & 4.80 & 3.37\ensuremath{\times} & 4.34 & 3.67\ensuremath{\times} & 4.72 & 2.39\ensuremath{\times} & 2.97 & 3.71\ensuremath{\times} & 4.81 \\
 & RKL & 2.95\ensuremath{\times} & 3.82 & 2.80\ensuremath{\times} & 3.63 & 2.34\ensuremath{\times} & 3.01 & 2.33\ensuremath{\times} & 3.03 & 2.34\ensuremath{\times} & 3.01 & 2.44\ensuremath{\times} & 3.11 & 1.71\ensuremath{\times} & 2.12 & 2.42\ensuremath{\times} & 3.10 \\
 & TV & 0.81\ensuremath{\times} & 1.02 & 0.81\ensuremath{\times} & 1.01 & 0.80\ensuremath{\times} & 1.01 & 0.80\ensuremath{\times} & 1.00 & 0.79\ensuremath{\times} & 1.01 & 0.82\ensuremath{\times} & 1.01 & 0.82\ensuremath{\times} & 1.01 & 0.81\ensuremath{\times} & 1.01 \\
 & LK & 4.90\ensuremath{\times} & 6.43 & 4.38\ensuremath{\times} & 5.77 & 3.58\ensuremath{\times} & 4.66 & 3.64\ensuremath{\times} & 4.80 & 3.37\ensuremath{\times} & 4.36 & 3.70\ensuremath{\times} & 4.77 & 2.38\ensuremath{\times} & 2.97 & 3.71\ensuremath{\times} & 4.82 \\
 & \textbf{BV (Ours)} & \textbf{5.67\ensuremath{\times}} & \textbf{7.18} & \textbf{4.98\ensuremath{\times}} & \textbf{6.36} & \textbf{4.12\ensuremath{\times}} & \textbf{5.27} & \textbf{4.07\ensuremath{\times}} & \textbf{5.17} & \textbf{3.71\ensuremath{\times}} & \textbf{4.71} & \textbf{3.96\ensuremath{\times}} & \textbf{5.00} & \textbf{2.56\ensuremath{\times}} & \textbf{3.21} & \textbf{4.15\ensuremath{\times}} & \textbf{5.27} \\
\cmidrule(lr){1-18}
\multirow{6}{*}{DSpark} & CE & 4.97\ensuremath{\times} & 6.70 & 4.61\ensuremath{\times} & 6.21 & 3.76\ensuremath{\times} & 5.02 & 3.80\ensuremath{\times} & 5.11 & 3.45\ensuremath{\times} & 4.60 & 3.75\ensuremath{\times} & 4.96 & 2.43\ensuremath{\times} & 3.10 & 3.82\ensuremath{\times} & 5.10 \\
 & KL & 5.33\ensuremath{\times} & 7.32 & 4.90\ensuremath{\times} & 6.83 & 4.11\ensuremath{\times} & 5.58 & 4.10\ensuremath{\times} & \textbf{5.57} & 3.64\ensuremath{\times} & 5.08 & 3.90\ensuremath{\times} & 5.31 & 2.62\ensuremath{\times} & 3.34 & 4.09\ensuremath{\times} & 5.58 \\
 & RKL & 3.20\ensuremath{\times} & 4.22 & 3.14\ensuremath{\times} & 4.13 & 2.75\ensuremath{\times} & 3.59 & 2.62\ensuremath{\times} & 3.44 & 2.48\ensuremath{\times} & 3.23 & 2.78\ensuremath{\times} & 3.60 & 1.82\ensuremath{\times} & 2.31 & 2.69\ensuremath{\times} & 3.50 \\
 & TV & 0.80\ensuremath{\times} & 1.02 & 0.80\ensuremath{\times} & 1.02 & 0.79\ensuremath{\times} & 1.01 & 0.80\ensuremath{\times} & 1.02 & 0.80\ensuremath{\times} & 1.02 & 0.82\ensuremath{\times} & 1.04 & 0.81\ensuremath{\times} & 1.01 & 0.80\ensuremath{\times} & 1.02 \\
 & LK & 5.46\ensuremath{\times} & 7.44 & 5.02\ensuremath{\times} & 6.79 & 4.10\ensuremath{\times} & 5.47 & 4.15\ensuremath{\times} & 5.55 & 3.85\ensuremath{\times} & 5.06 & 4.07\ensuremath{\times} & 5.36 & 2.60\ensuremath{\times} & 3.34 & 4.18\ensuremath{\times} & 5.57 \\
 & \textbf{BV (Ours)} & \textbf{5.78\ensuremath{\times}} & \textbf{7.64} & \textbf{5.29\ensuremath{\times}} & \textbf{6.95} & \textbf{4.41\ensuremath{\times}} & \textbf{5.69} & \textbf{4.32\ensuremath{\times}} & 5.56 & \textbf{3.99\ensuremath{\times}} & \textbf{5.09} & \textbf{4.23\ensuremath{\times}} & \textbf{5.38} & \textbf{2.63\ensuremath{\times}} & \textbf{3.37} & \textbf{4.38\ensuremath{\times}} & \textbf{5.67} \\
\midrule
\multicolumn{2}{@{}l@{\hspace{5pt}}}{$T=1$, token verify} & {\fontsize{6}{7}\selectfont Speedup} & $\tau$ & {\fontsize{6}{7}\selectfont Speedup} & $\tau$ & {\fontsize{6}{7}\selectfont Speedup} & $\tau$ & {\fontsize{6}{7}\selectfont Speedup} & $\tau$ & {\fontsize{6}{7}\selectfont Speedup} & $\tau$ & {\fontsize{6}{7}\selectfont Speedup} & $\tau$ & {\fontsize{6}{7}\selectfont Speedup} & $\tau$ & {\fontsize{6}{7}\selectfont Speedup} & $\tau$ \\
\midrule
\multirow{6}{*}{DFlash} & CE & 4.23\ensuremath{\times} & 5.31 & 3.74\ensuremath{\times} & 4.66 & 2.95\ensuremath{\times} & 3.65 & 3.31\ensuremath{\times} & 4.12 & 3.03\ensuremath{\times} & 3.75 & 3.39\ensuremath{\times} & 4.16 & 2.11\ensuremath{\times} & 2.62 & 3.25\ensuremath{\times} & 4.04 \\
 & KL & 4.26\ensuremath{\times} & 5.44 & 3.75\ensuremath{\times} & 4.76 & 3.07\ensuremath{\times} & 3.89 & 3.30\ensuremath{\times} & 4.18 & 3.00\ensuremath{\times} & 3.78 & 3.36\ensuremath{\times} & 4.20 & 2.12\ensuremath{\times} & 2.65 & 3.27\ensuremath{\times} & 4.13 \\
 & RKL & 2.98\ensuremath{\times} & 3.69 & 2.80\ensuremath{\times} & 3.46 & 2.29\ensuremath{\times} & 2.82 & 2.34\ensuremath{\times} & 2.89 & 2.38\ensuremath{\times} & 2.92 & 2.58\ensuremath{\times} & 3.13 & 1.68\ensuremath{\times} & 2.08 & 2.44\ensuremath{\times} & 3.00 \\
 & TV & 0.84\ensuremath{\times} & 1.02 & 0.84\ensuremath{\times} & 1.01 & 0.84\ensuremath{\times} & 1.01 & 0.83\ensuremath{\times} & 1.00 & 0.84\ensuremath{\times} & 1.01 & 0.84\ensuremath{\times} & 1.01 & 0.82\ensuremath{\times} & 1.01 & 0.83\ensuremath{\times} & 1.01 \\
 & LK & 4.39\ensuremath{\times} & 5.56 & 3.85\ensuremath{\times} & 4.85 & 3.12\ensuremath{\times} & 3.87 & 3.33\ensuremath{\times} & 4.20 & 3.11\ensuremath{\times} & 3.86 & 3.47\ensuremath{\times} & 4.30 & 2.17\ensuremath{\times} & 2.71 & 3.35\ensuremath{\times} & 4.19 \\
 & \textbf{BV (Ours)} & \textbf{5.05\ensuremath{\times}} & \textbf{6.53} & \textbf{4.43\ensuremath{\times}} & \textbf{5.68} & \textbf{3.60\ensuremath{\times}} & \textbf{4.60} & \textbf{3.76\ensuremath{\times}} & \textbf{4.82} & \textbf{3.48\ensuremath{\times}} & \textbf{4.43} & \textbf{3.83\ensuremath{\times}} & \textbf{4.90} & \textbf{2.43\ensuremath{\times}} & \textbf{3.05} & \textbf{3.80\ensuremath{\times}} & \textbf{4.86} \\
\cmidrule(lr){1-18}
\multirow{6}{*}{DSpark} & CE & 4.63\ensuremath{\times} & 6.14 & 4.18\ensuremath{\times} & 5.52 & 3.46\ensuremath{\times} & 4.55 & 3.48\ensuremath{\times} & 4.58 & 3.23\ensuremath{\times} & 4.24 & 3.43\ensuremath{\times} & 4.47 & 2.17\ensuremath{\times} & 2.90 & 3.51\ensuremath{\times} & 4.63 \\
 & KL & 4.74\ensuremath{\times} & 6.53 & 4.32\ensuremath{\times} & 5.91 & 3.45\ensuremath{\times} & 4.70 & 3.58\ensuremath{\times} & 4.92 & 3.32\ensuremath{\times} & 4.50 & 3.51\ensuremath{\times} & 4.71 & 2.26\ensuremath{\times} & 3.03 & 3.60\ensuremath{\times} & 4.90 \\
 & RKL & 3.03\ensuremath{\times} & 4.09 & 2.89\ensuremath{\times} & 3.89 & 2.53\ensuremath{\times} & 3.39 & 2.44\ensuremath{\times} & 3.30 & 2.34\ensuremath{\times} & 3.13 & 2.72\ensuremath{\times} & 3.59 & 1.68\ensuremath{\times} & 2.24 & 2.52\ensuremath{\times} & 3.38 \\
 & TV & 0.76\ensuremath{\times} & 1.02 & 0.77\ensuremath{\times} & 1.02 & 0.76\ensuremath{\times} & 1.01 & 0.76\ensuremath{\times} & 1.02 & 0.76\ensuremath{\times} & 1.02 & 0.79\ensuremath{\times} & 1.04 & 0.77\ensuremath{\times} & 1.01 & 0.77\ensuremath{\times} & 1.02 \\
 & LK & 4.97\ensuremath{\times} & 6.79 & 4.47\ensuremath{\times} & 6.08 & 3.44\ensuremath{\times} & 4.66 & 3.77\ensuremath{\times} & 5.11 & 3.46\ensuremath{\times} & 4.65 & 3.72\ensuremath{\times} & 4.96 & 2.32\ensuremath{\times} & 3.11 & 3.74\ensuremath{\times} & 5.05 \\
 & \textbf{BV (Ours)} & \textbf{5.18\ensuremath{\times}} & \textbf{7.13} & \textbf{4.57\ensuremath{\times}} & \textbf{6.27} & \textbf{3.74\ensuremath{\times}} & \textbf{5.14} & \textbf{3.83\ensuremath{\times}} & \textbf{5.18} & \textbf{3.55\ensuremath{\times}} & \textbf{4.77} & \textbf{3.80\ensuremath{\times}} & \textbf{5.13} & \textbf{2.38\ensuremath{\times}} & \textbf{3.19} & \textbf{3.86\ensuremath{\times}} & \textbf{5.26} \\
\midrule
\multicolumn{2}{@{}l@{\hspace{5pt}}}{$T=1$, block verify} & {\fontsize{6}{7}\selectfont Speedup} & $\tau$ & {\fontsize{6}{7}\selectfont Speedup} & $\tau$ & {\fontsize{6}{7}\selectfont Speedup} & $\tau$ & {\fontsize{6}{7}\selectfont Speedup} & $\tau$ & {\fontsize{6}{7}\selectfont Speedup} & $\tau$ & {\fontsize{6}{7}\selectfont Speedup} & $\tau$ & {\fontsize{6}{7}\selectfont Speedup} & $\tau$ & {\fontsize{6}{7}\selectfont Speedup} & $\tau$ \\
\midrule
\multirow{6}{*}{DFlash} & CE & 4.29\ensuremath{\times} & 5.49 & 3.77\ensuremath{\times} & 4.80 & 3.10\ensuremath{\times} & 3.91 & 3.32\ensuremath{\times} & 4.21 & 3.03\ensuremath{\times} & 3.81 & 3.36\ensuremath{\times} & 4.18 & 2.13\ensuremath{\times} & 2.67 & 3.29\ensuremath{\times} & 4.15 \\
 & KL & 4.38\ensuremath{\times} & 5.52 & 3.90\ensuremath{\times} & 4.89 & 3.29\ensuremath{\times} & 4.07 & 3.35\ensuremath{\times} & 4.20 & 3.08\ensuremath{\times} & 3.82 & 3.39\ensuremath{\times} & 4.17 & 2.13\ensuremath{\times} & 2.67 & 3.36\ensuremath{\times} & 4.19 \\
 & RKL & 2.95\ensuremath{\times} & 3.68 & 2.78\ensuremath{\times} & 3.46 & 2.36\ensuremath{\times} & 2.91 & 2.35\ensuremath{\times} & 2.95 & 2.40\ensuremath{\times} & 2.94 & 2.57\ensuremath{\times} & 3.11 & 1.66\ensuremath{\times} & 2.07 & 2.44\ensuremath{\times} & 3.02 \\
 & TV & 0.82\ensuremath{\times} & 1.02 & 0.81\ensuremath{\times} & 1.01 & 0.82\ensuremath{\times} & 1.01 & 0.81\ensuremath{\times} & 1.00 & 0.81\ensuremath{\times} & 1.01 & 0.82\ensuremath{\times} & 1.01 & 0.81\ensuremath{\times} & 1.01 & 0.81\ensuremath{\times} & 1.01 \\
 & LK & 4.38\ensuremath{\times} & 5.70 & 3.92\ensuremath{\times} & 4.98 & 3.14\ensuremath{\times} & 3.92 & 3.39\ensuremath{\times} & 4.27 & 3.14\ensuremath{\times} & 3.96 & 3.51\ensuremath{\times} & 4.36 & 2.21\ensuremath{\times} & 2.79 & 3.38\ensuremath{\times} & 4.28 \\
 & \textbf{BV (Ours)} & \textbf{5.28\ensuremath{\times}} & \textbf{6.79} & \textbf{4.46\ensuremath{\times}} & \textbf{5.82} & \textbf{3.82\ensuremath{\times}} & \textbf{4.89} & \textbf{3.83\ensuremath{\times}} & \textbf{4.91} & \textbf{3.49\ensuremath{\times}} & \textbf{4.47} & \textbf{3.85\ensuremath{\times}} & \textbf{4.90} & \textbf{2.45\ensuremath{\times}} & \textbf{3.08} & \textbf{3.88\ensuremath{\times}} & \textbf{4.98} \\
\cmidrule(lr){1-18}
\multirow{6}{*}{DSpark} & CE & 4.66\ensuremath{\times} & 6.38 & 4.16\ensuremath{\times} & 5.65 & 3.32\ensuremath{\times} & 4.49 & 3.45\ensuremath{\times} & 4.70 & 3.23\ensuremath{\times} & 4.35 & 3.37\ensuremath{\times} & 4.50 & 2.18\ensuremath{\times} & 2.93 & 3.48\ensuremath{\times} & 4.71 \\
 & KL & 4.94\ensuremath{\times} & 6.75 & 4.50\ensuremath{\times} & 6.11 & 3.66\ensuremath{\times} & 4.93 & 3.77\ensuremath{\times} & 5.13 & 3.47\ensuremath{\times} & 4.65 & 3.62\ensuremath{\times} & 4.80 & 2.34\ensuremath{\times} & 3.14 & 3.76\ensuremath{\times} & 5.07 \\
 & RKL & 3.03\ensuremath{\times} & 4.10 & 2.90\ensuremath{\times} & 3.93 & 2.51\ensuremath{\times} & 3.40 & 2.49\ensuremath{\times} & 3.38 & 2.35\ensuremath{\times} & 3.16 & 2.67\ensuremath{\times} & 3.55 & 1.69\ensuremath{\times} & 2.27 & 2.52\ensuremath{\times} & 3.40 \\
 & TV & 0.77\ensuremath{\times} & 1.02 & 0.77\ensuremath{\times} & 1.02 & 0.77\ensuremath{\times} & 1.01 & 0.77\ensuremath{\times} & 1.02 & 0.77\ensuremath{\times} & 1.02 & 0.80\ensuremath{\times} & 1.04 & 0.76\ensuremath{\times} & 1.01 & 0.77\ensuremath{\times} & 1.02 \\
 & LK & 5.12\ensuremath{\times} & 7.07 & 4.62\ensuremath{\times} & 6.35 & 3.68\ensuremath{\times} & 4.98 & 3.73\ensuremath{\times} & 5.11 & 3.48\ensuremath{\times} & 4.71 & 3.70\ensuremath{\times} & 4.98 & 2.38\ensuremath{\times} & 3.20 & 3.82\ensuremath{\times} & 5.20 \\
 & \textbf{BV (Ours)} & \textbf{5.34\ensuremath{\times}} & \textbf{7.33} & \textbf{4.72\ensuremath{\times}} & \textbf{6.49} & \textbf{3.95\ensuremath{\times}} & \textbf{5.33} & \textbf{4.04\ensuremath{\times}} & \textbf{5.47} & \textbf{3.67\ensuremath{\times}} & \textbf{4.92} & \textbf{3.87\ensuremath{\times}} & \textbf{5.19} & \textbf{2.41\ensuremath{\times}} & \textbf{3.25} & \textbf{4.00\ensuremath{\times}} & \textbf{5.43} \\
\bottomrule
\end{tabular}
}
\end{table}

\end{document}